\documentclass[10pt,twocolumn,letterpaper]{article}

\usepackage{wacv}              

\usepackage{graphicx}
\usepackage{amsmath}
\usepackage{amssymb}
\usepackage{booktabs}
\usepackage{xcolor}
\usepackage{multirow} 

\usepackage{tabularx}
\newcolumntype{Y}{>{\centering\arraybackslash}X}

\usepackage[pagebackref,breaklinks,colorlinks]{hyperref}

\newcommand\blfootnote[1]{%
  \begingroup
  \renewcommand\thefootnote{}\footnote{#1}%
  \addtocounter{footnote}{-1}%
  \endgroup
}

\usepackage[capitalize]{cleveref}
\crefname{section}{Sec.}{Secs.}
\Crefname{section}{Section}{Sections}
\Crefname{table}{Table}{Tables}
\crefname{table}{Tab.}{Tabs.}

\def\wacvPaperID{1797} 
\def\confName{WACV}
\def\confYear{2024}

\begin{document}

\title{Diffusion models meet image counter-forensics}

\author{Matías Tailanián$^{\dagger,1,2}$, Marina Gardella$^{\dagger,3}$, Alvaro Pardo$^{2,4}$, Pablo Musé$^{1}$\\
$^\dagger$Equally contributing authors.\\
$^1$IIE, Facultad de Ingeniería,Universidad de la República\\
$^2$Digital Sense\\
$^3$Centre Borelli, École Normale Supérieure Paris-Saclay, Université Paris-Saclay\\
$^4$Universidad Católica del Uruguay\\
}
\maketitle

\begin{abstract}
From its acquisition in the camera sensors to its storage, different operations are performed to generate the final image. This pipeline imprints specific traces into the image to form a natural watermark. Tampering with an image disturbs these traces; these disruptions are clues that are used by most methods to detect and locate forgeries.
In this article, we assess the capabilities of diffusion models to erase the traces left by forgers and, therefore, deceive forensics methods. Such an approach has been recently introduced for adversarial purification, achieving significant performance. We show that diffusion purification methods are well suited for counter-forensics tasks. Such approaches outperform already existing counter-forensics techniques both in deceiving forensics methods and in preserving the natural look of the purified images. 
The source code is publicly available at \url{https://github.com/mtailanian/diff-cf}.
\end{abstract}
\blfootnote{This work has received funding by the Paris Region Ph.D. grant from Région Île-de-France, the ANR project APATE (ANR-22-CE39-0016), the European Union under the Horizon Europe VERA.AI project, Grant Agreement number 101070093 and by a graduate scholarship from Agencia Nacional de Investigación e Innovación, Uruguay.}

\vspace*{-0.5 cm}
\section{Introduction}
\label{sec:intro}

Image forgeries are present everywhere~\cite{farid_book}, from fake news on social media~\cite{qureshi2014review} to scientific misconduct. Indeed, many image processing tools are available to create visually realistic image alterations. Yet, these modifications leave traces on the image that are tampering cues. Image forensics aims at detecting these alterations by finding local inconsistencies~\cite{farid_book}. Image counter-forensics emerged as the research field that challenges forensics methods and explores their limitations~\cite{bohme2013counter}.

Adversarial attacks share some common properties with image forgeries in the sense that both techniques introduce subtle modifications to the images that, though imperceptible to the naked eye, disrupt the image's traces. The goal of adversarial attacks is to deceive a model into making incorrect predictions. Adversarial purification can be, therefore, linked to counter-forensics since it aims at preprocessing the input data to remove these adversarial perturbations. Generally, these purification methods are based on generative models~\cite{samangouei2018defensegan}.

In recent years, diffusion models have emerged as highly effective generative models~\cite{NEURIPS2020_DDPM, song2021scorebased}. These models have showcased impressive capabilities in generating high-quality samples, outperforming traditional Generative Adversarial Networks (GANs) in the realm of image generation. The advancements in diffusion models have led to significant improvements in the fidelity and realism of synthesized images, highlighting their potential as state-of-the-art models in the field.

\begin{figure}[t]
    \centering
    \includegraphics[width= \columnwidth]{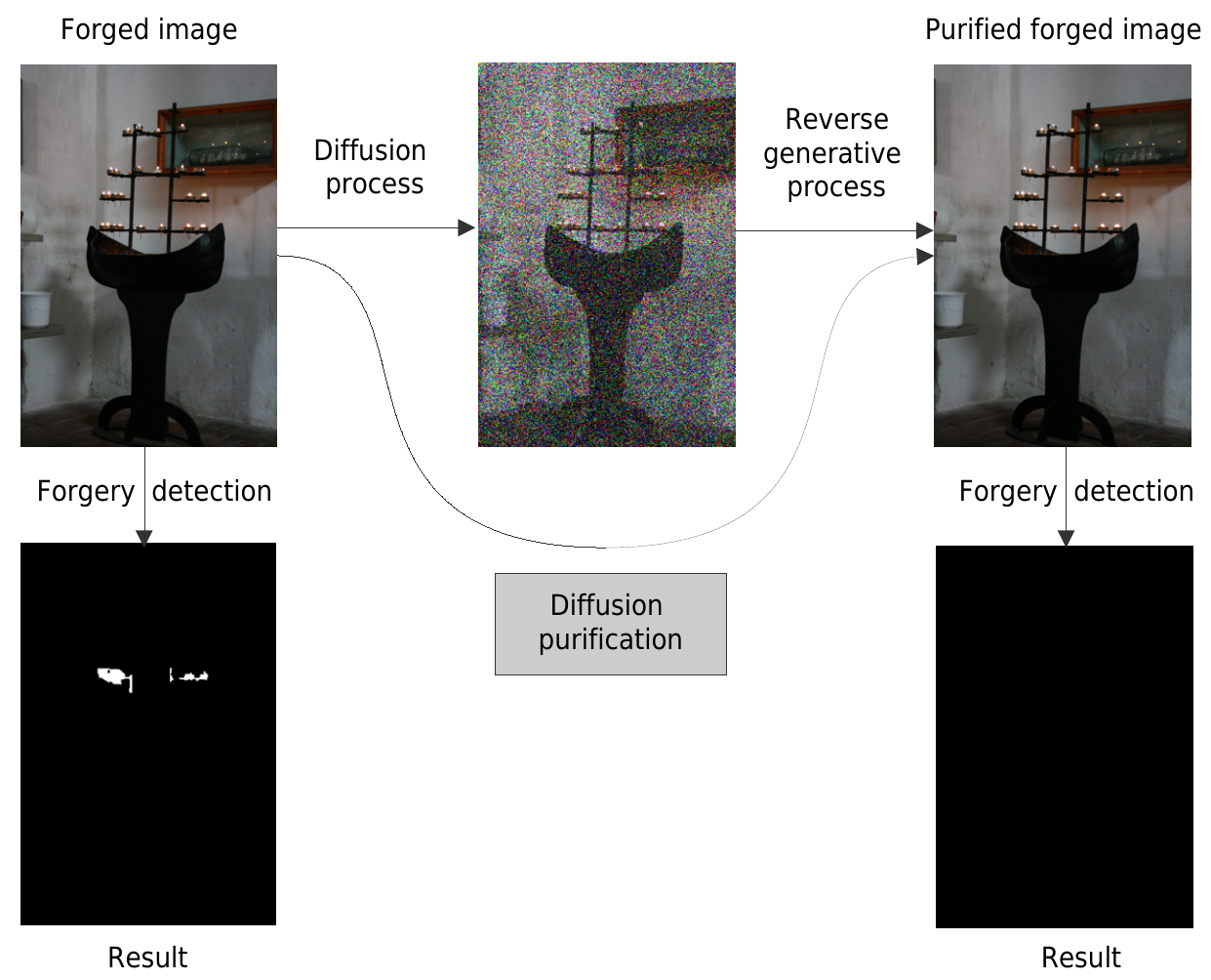}
    \caption{Illustration of using diffusion models as a counter-forensic technique. A forged image from FAU dataset~\cite{FAU_dataset}, correctly detected by ZERO~\cite{ZERO}, produces no detection after diffusion purification.}
\label{fig:teaser}
\vspace*{-0.3 cm}
\end{figure}

In this work, we evaluate, for the first time, the efficiency of diffusion purification methods, currently used for adversarial purification~\cite{nie2022DiffPure, GuidedDiffAdvPurif}, as counter-forensics methods. The rationale behind the use of diffusion models for adversarial purification is that these models learn the distribution of clean data. Hence, by diffusing an adversarial example and then applying the reverse generative process, the diffusion model gradually removes the adversarial perturbations and reconstructs the underlying clean sample.

The same rationale can be applied to hide the forensic traces caused by tampering. Indeed, since diffusion models are trained on pristine images, diffusion purification methods applied to forged images should recover purified images without any inconsistency in the camera traces. Once such disruptions on the camera processing chain are erased, purified images should be able to deceive any forgery detection method relying on them. Fig.~\ref{fig:teaser} shows an example of the aforementioned approach: while ZERO~\cite{ZERO} correctly detects the original forgery, once diffusion purification is applied, the method is no longer able to detect it. 

\section{Related work}

\subsection{Image counter-forensics}

Counter-forensics attacks can be classified into two categories: the first one corresponds to those that focus on a specific trace or method, while the second category corresponds to generic attacks that aim at erasing all the forgery traces and should, therefore, be able to deceive any forensic method.
Among methods of the first category, Fan~et~al.~\cite{fan2013jpeg} and Comesana~et~al.~\cite{comesana2014optimal} propose attacks against histogram-based methods, mainly used to detect JPEG compression traces. Kirchner~et~al.~\cite{kirchner2008hiding} propose hiding resampling traces by removing the periodic variations in the residual signal in the spatial domain. Do~et~al.~\cite{do2010deluding} design SIFT-specific attacks that are able to deceive copy-move forgery detectors based on such local descriptors. 

With the advent of learning-based forgery detectors, counter-forensic attacks specifically designed for such methods have also been proposed. Marra~et~al.~\cite{marra2015counter} design a counter-forensic scheme on the feature space. Their goal is to restore the features of the pristine image and, by doing so, to cross the decision boundary of the target detector. In the case of perfect knowledge of the target method, this counter-forensic method delivers great results. However, when the target detector is unknown, the results degrade tremendously. Other methods countering specific learning-based detectors with an optimum attack which relies on gradient descent solutions have also been explored~\cite{chen2017gradient, gragnaniello2018analysis}.

With limited knowledge of forensic models, counter-forensics attackers focus more on erasing the traces by generic tools~\cite{barni2018adversarial}.
The median filter is a technique commonly used as an anti-forensics attack~\cite{wu2013anti}, in deep convolutional neural network versions~\cite{kim2017median} or even variational formulations~\cite{singh2019improved}. Though this method can be effective on several traces, it leaves a distinctive streaking artifact that can be retrieved~\cite{kirchnner-median-filtering, forensics-median-filtering}. To compensate for this, techniques to remove such artifacts have been proposed~\cite{fontani2012hiding, singh2019improved}.

More recently, Chen~et~al.~\cite{Chen_2020_CVPR} proposed erasing camera traces, trying not to damage the signal content by adopting a Siamese-based neural network. Cozzolino~et~al.~\cite{Cozzolino_2021_CVPR} and Wu~et~al.~\cite{wu2019multiple} use generative adversarial approaches. Baracchi~et~al.~\cite{BARACCHI2021301213} exploit a real camera firmware to perform the manipulation while reproducing the image statistics. This approach can be most efficient at creating real camera traces, and can easily fool camera identification methods into thinking the image was taken with this camera. However, this method is difficult to use, since it requires disassembling a camera to hack its input field.

\subsection{Diffusion-based adversarial purification}

Nie et al.~\cite{nie2022DiffPure} were the first ones to propose the use of the forward and reverse processes of a pre-trained diffusion model for image adversarial purification. Their method --DiffPure-- first diffuses adversarial examples with a small amount of noise. Then, the clean image is recovered through the reverse generative process. A very similar idea was developed at the same time by Blau et al.~\cite{blau2022threat}.
The theoretical fundamentals justifying the performance of such diffusion-based adversarial purification methods are derived in~\cite{DensePure}.

Wang et al.~\cite{GuidedDiffAdvPurif} face the difficult trade-off between choosing a long diffusion time, which guarantees the removal of the adversarial perturbation, and choosing a small one, which guarantees the similarity between the input image and the purified one. They propose to guide the reverse process by the adversarial image. By doing so, the purified image is forced to stay close to the input image.

Wu et al.~\cite{GuidedDiffAdvPurifFROM_RANDOM_NOISE} also guide the reverse process by the adversarial image. However, they propose to sample the initial input from pure Gaussian noise and gradually denoise it. The rationale of their approach is that the diffused image still carries corrupted structures, and the reverse process is likely to get stuck in those corrupted structures.

As the field evolves, several applications of these approaches have been developed. In~\cite{DenoisingDiffusionProbabilisticModelsDefenseAdversarialAttacks}, the authors analyze the performance of DiffPure~\cite{nie2022DiffPure} to purify adversarial attacks on the classification of metastatic tissue. In~\cite{Adversarial3Dpointcloud}, the authors apply the same principle as in~\cite{nie2022DiffPure} but using an extension of diffusion models to the 3D space~\cite{Luo_2021_CVPR}. Similarly,~\cite{AdversarialAudio} also shares the grounds of DiffPure~\cite{nie2022DiffPure} but using a waveform-based diffusion model~\cite{kong2021diffwave} for adversarial audio purification.

Diffusion purification methods have rapidly gained attention in the field. This interest has even led to questioning the evaluation practices of such techniques~\cite{EvalDiffusionPurif}.


\section{Background}

In this section, we provide a brief overview of Denoising Diffusion Models~\cite{NEURIPS2020_DDPM, pmlr-v37-sohl-dickstein15, song2021scorebased} that will be used as a basis for the next section. Recently, denoising diffusion models, alternatively called score-based generative models, have emerged as a powerful approach amongst generative methods. Denoising diffusion models consist of two processes: a forward diffusion process that progressively adds noise to the input and a reverse generative process that learns to generate data by denoising.

\paragraph{Forward diffusion process.} The diffusion process is a Markov process that gradually adds noise to the clean input data. Let $T$ be the number of steps of the diffusion process, $\mathbf{x}_0$ an input image, and $\mathbf{x}_t$ the forward image until step $t$ ($0 \leq t \leq T$). The diffusion process from clean data $\mathbf{x}_0$ to $\mathbf{x}_T$ is defined as
\begin{equation}
q(\mathbf{x}_{1:T}| \mathbf{x}_0) = \prod_{t=1}^T q(\mathbf{x}_t|\mathbf{x}_{t-1}),
\end{equation}
\begin{equation}
    \text{with }q(\mathbf{x}_t|\mathbf{x}_{t-1}) = \mathcal{N}(\mathbf{x}_t; \sqrt{1-\beta_t} \mathbf{x}_{t-1}, \beta_t \mathbf{I}),
\end{equation}
\noindent where the variances $\beta_1, \dots, \beta_T$ are predefined small values.

A notable characteristic of the forward process is that there is a closed-form to generate $\mathbf{x_t}$ at any given time step $t$ directly from $\mathbf{x_0}$~\cite{NEURIPS2020_DDPM}. Indeed, let $\bar{\alpha}_t = \prod_{s=1}^t(1-\beta_s)$, then 
we can directly sample $\mathbf{x}_t$ as
\begin{equation}
    \mathbf{x}_t = \sqrt{\bar{\alpha}_t}\, \mathbf{x}_0 + \sqrt{(1-\bar{\alpha})}\, \epsilon , \;\text{where}\; \epsilon \sim \mathcal{N}(\mathbf{0}, \mathbf{I}).
\end{equation}

\paragraph{Reverse denoising process.} The reverse generative process is a Markov process that gradually eliminates the noise added in the forward process. The reverse process from $\mathbf{x}_T$ to $\mathbf{x}_0$ is given by
\begin{equation}
    p_\theta(\mathbf{x}_{0:T-1}| \mathbf{x}_T) = \prod_{t=1}^T p(\mathbf{x}_{t-1}|\mathbf{x}_{t})
\end{equation}
%
%
\begin{equation}
 \text{with }   p_\theta(\mathbf{x}_{t-1}|\mathbf{x}_{t}) = \mathcal{N}(\mathbf{x}_{t-1}; \mu_\theta(\mathbf{x}_t, t), \sigma^2_t I),
    \label{ec:p_theta_naive}
\end{equation}
\noindent where the mean $\mu_\theta(\mathbf{x}_t, t)$ is a trainable network and the variances $\sigma^2_0, \dots, \sigma_T^2$ can either be fixed or learned using a neural network.

\section{Proposed method}


Our goal is to introduce subtle modifications to a forged image to erase the traces left by the tampering process while, at the same time, preserving the semantic content. Our proposed approach is based on diffusion purification methods~\cite{GuidedDiffAdvPurif, GuidedDiffAdvPurifFROM_RANDOM_NOISE, nie2022DiffPure}. It consists of two steps: first, we add noise up to a certain time-step $t=t^*$ in the forward diffusion process, and then we gradually remove it following the reverse diffusion process, up to $t=0$. We refer to this method as \emph{Diffusion Counter-Forensics}, or shortly \emph{Diff-CF}.

The intuition behind this idea is that the probability distributions of the forged and its corresponding clean image are separated in $t=0$, but by adding noise in the forward process, the boundaries between the distributions get fuzzier, and they begin to overlap, more so the higher the value of $t^*$. Then, starting from a noisy sample that can belong to either probability distribution, the reverse diffusion process, which was trained on pristine images only, generates a purified version of the image with no forgery traces. See, for instance, Fig. 2 in~\cite{meng2021sdedit}.


The value $t^*$ plays a fundamental role. Intuitively, $t^*$ has to be large enough so that the noise added hides the forgery traces, but small enough so that we can preserve the image semantics and structure. If we set the value of $t^*$ too high, the resulting image would deviate too much from the original one. On the other hand, if the value of $t^*$ is too small, we might be unable to erase the forgery traces correctly. This trade-off is studied more in-depth in Sec.~\ref{sec:ablation-time}. 

With the purpose of being able to use larger values of $t^*$ without deviating too much from the input image, we also analyze the introduction of guidance in the reverse diffusion process. We refer to this variant as \emph{Counter-Forensics Guided Diffusion}, or \emph{Diff-CFG}.
More precisely, we propose to guide the reverse process using the forged image itself, as in~\cite{GuidedDiffAdvPurif}. In this way, we encourage the network to produce a clean image as close as possible to the forged one, under the assumption that the forgery traces are subtle enough that they are not reconstructed. In the normal reverse diffusion process, at each time step, a new image is sampled following Eq.~\ref{ec:p_theta_naive}. Instead, for this variant, we propose to sample from
\begin{equation} \label{eq:guided}
\begin{split}
    p_\theta( & \mathbf{x}_{t-1}| \mathbf{x}_{t}) = \\
    & = \mathcal{N}(\mathbf{x}_{t-1}; \mu_\theta(\mathbf{x}_t, t) - s_t\Sigma \nabla_{x_t} \mathcal{D}(x_t, x_{in}), \sigma^2_t I),
\end{split}
\end{equation}
where $\Sigma$ is the variance of $x_t$, $\mathcal{D}(x_t, x_{in})$ is some similarity measure between $x_t$ and the input image (forged image) $x_{in}$, and $s_t$ is a scale factor that depends on the time step $t$. For high values of $t$, the forgery traces are completely hidden by the added noise, so we can afford to use large values for $s_t$, without the risk of guiding the process to reconstruct the forged traces. On the other hand, for small values of $t$, the forgery traces are more retained, and therefore we should use smaller values for $s_t$. Similar to what is proposed in~\cite{GuidedDiffAdvPurif, GuidedDiffAdvPurifFROM_RANDOM_NOISE}, we define $s_t$ to be proportional to the added noise, as 
\begin{equation}
    s_t = s \, \frac{\sqrt{1 - \bar{\alpha}_t}}{\sqrt{\bar{\alpha}_t}},
    \label{ec:scale}
\end{equation}

where $s$ is a hyper-parameter. \\

\definecolor{c1}{HTML}{F6830F}
\definecolor{c2}{HTML}{edd3b8}
\definecolor{c3}{HTML}{1F3C88}
\definecolor{c4}{HTML}{c6cfea}

\newcommand{\mcc}[1]{\textcolor{c1}{\sffamily\fontsize{7.5pt}{7.5pt}\selectfont#1}}
\newcommand{\mccb}[1]{\textcolor{c1}{\sffamily\fontsize{7.5pt}{7.5pt}\selectfont\bfseries\underline{#1}}}
\newcommand{\gomcc}[1]{\textcolor{c2}{\sffamily\fontsize{7.5pt}{7.5pt}\selectfont#1}}
\newcommand{\iou}[1]{\textcolor{c3}{\sffamily\fontsize{7.5pt}{7.5pt}\selectfont#1}}
\newcommand{\ioub}[1]{\textcolor{c3}{\sffamily\fontsize{7.5pt}{7.5pt}\selectfont\bfseries\underline{#1}}}
\newcommand{\goiou}[1]{\textcolor{c4}{\sffamily\fontsize{7.5pt}{7.5pt}\selectfont#1}}


\begin{table*}[t]  
  \small  
  \centering
  \setlength{\tabcolsep}{0em}
  \begin{tabularx}{1\textwidth}{c@{\hskip 5pt}l@{\hskip 5pt}llllllllll}

\toprule

 & & CatNet & Choi & Comprint & MantraNet & Noiseprint & Shin & SpliceBuster & TruFor & ZERO & Avg$_w$\\ 
\midrule

\multirow{10}{*}[.4em]{\rotatebox[origin=c]{90}{\footnotesize{Korus  }}} & \multirow{2}{*}{\rotatebox[origin=c]{90}{\tiny{Original}}} &\mcc{ 0.0790  }&\mcc{ 0.1971  }&\mcc{ 0.0534  }&\mcc{ 0.1261  }&\mcc{ 0.0988  }&\gomcc{ 0.0221  }&\mcc{ 0.1405  }&\mcc{ 0.3428  } & \gomcc{0.0050} & \mcc{-} \\
& &\iou{ 0.0433  }&\iou{ 0.1261  }&\iou{ 0.0461  }&\iou{ 0.0982  }&\iou{ 0.0792  }&\goiou{ 0.0568  }&\iou{ 0.1012  }&\iou{ 0.2575  } & \goiou{0.0028} & \iou{-} \\[3pt]

& \multirow{2}{*}{\rotatebox[origin=c]{90}{\tiny{CamTE}}} &\mccb{ 0.0468 \tiny{(-0.0322)} }&\mcc{ 0.0597 \tiny{(-0.1374)} }&\mcc{ 0.0356 \tiny{(-0.0179)} }&\mcc{ 0.0646 \tiny{(-0.0614)} }&\mcc{ 0.0420 \tiny{(-0.0569)} }&\gomcc{ 0.0305 \tiny{(0.0084)} }&\mcc{ 0.0817 \tiny{(-0.0588)} }&\mccb{ 0.1961 \tiny{(-0.1467)} } & \gomcc{0.0000 \tiny{(-0.0050)}} & \mcc{-0.1024}\\
& &\ioub{ 0.0278 \tiny{(-0.0155)} }&\iou{ 0.0400 \tiny{(-0.0860)} }&\iou{ 0.0389 \tiny{(-0.0072)} }&\iou{ 0.0644 \tiny{(-0.0338)} }&\iou{ 0.0545 \tiny{(-0.0247)} }&\goiou{ 0.0578 \tiny{(0.0011)} }&\iou{ 0.0729 \tiny{(-0.0284)} }&\ioub{ 0.1489 \tiny{(-0.1086)} } & \goiou{0.0000 \tiny{(-0.0028)}} & \iou{-0.0479}\\[5pt]

& \multirow{2}{*}{\rotatebox[origin=c]{90}{\tiny{BM3D}}} &\mcc{ 0.0997 \tiny{(0.0207)} }&\mcc{ 0.0352 \tiny{(-0.1619)} }&\mcc{ 0.0278 \tiny{(-0.0256)} }&\mcc{ 0.0652 \tiny{(-0.0609)} }&\mcc{ 0.0420 \tiny{(-0.0569)} }&\gomcc{ 0.0155 \tiny{(-0.0066)} }&\mcc{ 0.0860 \tiny{(-0.0545)} }&\mcc{ 0.2579 \tiny{(-0.0850)} } & \gomcc{0.0000 \tiny{(-0.0050)}} & \mcc{-0.0819}\\
& &\iou{ 0.0646 \tiny{(0.0213)} }&\ioub{ 0.0227 \tiny{(-0.1033)} }&\iou{ 0.0346 \tiny{(-0.0115)} }&\iou{ 0.0746 \tiny{(-0.0237)} }&\iou{ 0.0514 \tiny{(-0.0278)} }&\goiou{ 0.0540 \tiny{(-0.0028)} }&\iou{ 0.0744 \tiny{(-0.0268)} }&\iou{ 0.1964 \tiny{(-0.0611)} } & \goiou{0.0000 \tiny{(-0.0028)}} & \iou{-0.0358}\\[5pt]

& \multirow{2}{*}{\rotatebox[origin=c]{90}{\tiny{Diff-CF}}} &\mccb{ 0.0418 \tiny{(-0.0371)} }&\mccb{ 0.0147 \tiny{(-0.1825)} }&\mccb{ 0.0024 \tiny{(-0.0510)} }&\mccb{ 0.0255 \tiny{(-0.1005)} }&\mccb{ 0.0190 \tiny{(-0.0798)} }&\gomcc{ 0.0027 \tiny{(-0.0194)} }&\mccb{ 0.0350 \tiny{(-0.1055)} }&\mccb{ 0.1454 \tiny{(-0.1974)} } & \gomcc{0.0045 \tiny{(-0.0005)}} & \mccb{-0.1451}\\
& &\ioub{ 0.0204 \tiny{(-0.0229)} }&\iou{ 0.0246 \tiny{(-0.1014)} }&\ioub{ 0.0215 \tiny{(-0.0246)} }&\ioub{ 0.0416 \tiny{(-0.0566)} }&\ioub{ 0.0360 \tiny{(-0.0432)} }&\goiou{ 0.0487 \tiny{(-0.0081)} }&\ioub{ 0.0401 \tiny{(-0.0611)} }&\ioub{ 0.1131 \tiny{(-0.1445)} } & \goiou{0.0027 \tiny{(-0.0001)}} & \ioub{-0.0677}\\[5pt]

& \multirow{2}{*}{\rotatebox[origin=c]{90}{\tiny{Diff-CFG}}} &\mcc{ 0.0852 \tiny{(0.0063)} }&\mccb{ 0.0044 \tiny{(-0.1927)} }&\mccb{ 0.0125 \tiny{(-0.0409)} }&\mccb{ 0.0442 \tiny{(-0.0818)} }&\mccb{ 0.0267 \tiny{(-0.0722)} }&\gomcc{ 0.0040 \tiny{(-0.0181)} }&\mccb{ 0.0456 \tiny{(-0.0950)} }&\mcc{ 0.2064 \tiny{(-0.1364)} } & \gomcc{0.0005 \tiny{(-0.0045)}} & \mccb{-0.1177} \\
& &\iou{ 0.0527 \tiny{(0.0095)} }&\ioub{ 0.0043 \tiny{(-0.1217)} }&\ioub{ 0.0281 \tiny{(-0.0180)} }&\ioub{ 0.0552 \tiny{(-0.0430)} }&\ioub{ 0.0446 \tiny{(-0.0346)} }&\goiou{ 0.0491 \tiny{(-0.0076)} }&\ioub{ 0.0488 \tiny{(-0.0525)} }&\iou{ 0.1601 \tiny{(-0.0975)} }& \goiou{0.0011 \tiny{(-0.0017)}} & \ioub{-0.0536}\\

\midrule

\multirow{10}{*}[.4em]{\rotatebox[origin=c]{90}{\footnotesize{FAU  }}} & \multirow{2}{*}{\rotatebox[origin=c]{90}{\tiny{Original}}} &\mcc{ 0.3228  }&\mcc{ 0.3045  }&\mcc{ 0.0393  }&\gomcc{ 0.0203  }&\mcc{ 0.0358  }&\mcc{ 0.1134  }&\gomcc{ 0.0074  }&\mcc{ 0.4039  }&\mcc{ 0.5855 } & \mcc{-}\\
& &\iou{ 0.2329  }&\iou{ 0.2670  }&\iou{ 0.0305  }&\goiou{ 0.0336  }&\iou{ 0.0482  }&\iou{ 0.1289  }&\goiou{ 0.0251  }&\iou{ 0.3373  }&\iou{ 0.5003} & \iou{-} \\[3pt]

& \multirow{2}{*}{\rotatebox[origin=c]{90}{\tiny{CamTE}}} &\mccb{ 0.0141 \tiny{(-0.3087)} }&\mcc{ 0.1426 \tiny{(-0.1620)} }&\mcc{ 0.0092 \tiny{(-0.0302)} }&\gomcc{ 0.0154 \tiny{(-0.0049)} }&\mcc{ 0.0242 \tiny{(-0.0116)} }&\mcc{ 0.0826 \tiny{(-0.0308)} }&\gomcc{ 0.0045 \tiny{(-0.0029)} }&\mcc{ 0.0553 \tiny{(-0.3486)} }&\mcc{ 0.0441 \tiny{(-0.5414)}} & \mcc{-0.6120}\\
& &\ioub{ 0.0085 \tiny{(-0.2244)} }&\iou{ 0.1173 \tiny{(-0.1497)} }&\iou{ 0.0206 \tiny{(-0.0099)} }&\goiou{ 0.0427 \tiny{(0.0091)} }&\iou{ 0.0434 \tiny{(-0.0049)} }&\iou{ 0.1046 \tiny{(-0.0243)} }&\goiou{ 0.0277 \tiny{(0.0026)} }&\ioub{ 0.0572 \tiny{(-0.2801)} }&\iou{ 0.0288 \tiny{(-0.4715)}} & \iou{-0.4259}\\[5pt]

& \multirow{2}{*}{\rotatebox[origin=c]{90}{\tiny{BM3D}}} &\mcc{ 0.0757 \tiny{(-0.2471)} }&\mcc{ 0.0679 \tiny{(-0.2367)} }&\mccb{ -0.0017 \tiny{(-0.0410)} }&\gomcc{ -0.0268 \tiny{(-0.0470)} }&\mccb{ -0.0014 \tiny{(-0.0372)} }&\mcc{ 0.0411 \tiny{(-0.0723)} }&\gomcc{ 0.0011 \tiny{(-0.0064)} }&\mcc{ 0.0802 \tiny{(-0.3237)} }&\mccb{ 0.0393 \tiny{(-0.5462)}} & \mcc{-0.6145}\\
& &\iou{ 0.0517 \tiny{(-0.1812)} }&\iou{ 0.0559 \tiny{(-0.2111)} }&\ioub{ 0.0126 \tiny{(-0.0179)} }&\goiou{ 0.0377 \tiny{(0.0041)} }&\iou{ 0.0331 \tiny{(-0.0152)} }&\iou{ 0.0803 \tiny{(-0.0486)} }&\goiou{ 0.0243 \tiny{(-0.0008)} }&\iou{ 0.0799 \tiny{(-0.2574)} }&\iou{ 0.0266 \tiny{(-0.4737)}} & \iou{-0.4298}\\[5pt]

& \multirow{2}{*}{\rotatebox[origin=c]{90}{\tiny{Diff-CF}}} &\mccb{ 0.0070 \tiny{(-0.3157)} }&\mccb{ 0.0242 \tiny{(-0.2803)} }&\mcc{ 0.0001 \tiny{(-0.0392)} }&\gomcc{ 0.0057 \tiny{(-0.0146)} }&\mccb{ -0.0018 \tiny{(-0.0376)} }&\mccb{ 0.0128 \tiny{(-0.1006)} }&\gomcc{ -0.0050 \tiny{(-0.0124)} }&\mccb{ 0.0399 \tiny{(-0.3640)} }&\mcc{ -0.0007 \tiny{(-0.5862)}} & \mccb{-0.6922}\\
& &\ioub{ 0.0056 \tiny{(-0.2273)} }&\ioub{ 0.0458 \tiny{(-0.2213)} }&\iou{ 0.0159 \tiny{(-0.0146)} }&\goiou{ 0.0355 \tiny{(0.0019)} }&\ioub{ 0.0199 \tiny{(-0.0283)} }&\ioub{ 0.0602 \tiny{(-0.0687)} }&\goiou{ 0.0123 \tiny{(-0.0128)} }&\ioub{ 0.0520 \tiny{(-0.2853)} }&\ioub{ 0.0015 \tiny{(-0.4988)}} & \ioub{-0.4687}\\[5pt]

& \multirow{2}{*}{\rotatebox[origin=c]{90}{\tiny{Diff-CFG}}} &\mcc{ 0.0241 \tiny{(-0.2986)} }&\mccb{ 0.0137 \tiny{(-0.2908)} }&\mccb{-0.0059 \tiny{(-0.0452)} }&\gomcc{ 0.0128 \tiny{(-0.0075)} }&\mcc{ 0.0002 \tiny{(-0.0355)} }&\mccb{ 0.0202 \tiny{(-0.0933)} }&\gomcc{ 0.0127 \tiny{(0.0053)} }&\mccb{ 0.0470 \tiny{(-0.3569)} }&\mccb{-0.0043 \tiny{(-0.5898)}} & \mccb{-0.6882}\\
& &\iou{ 0.0184 \tiny{(-0.2145)} }&\ioub{ 0.0220 \tiny{(-0.2451)} }&\ioub{ 0.0143 \tiny{(-0.0162)} }&\goiou{ 0.0339 \tiny{(0.0003)} }&\ioub{ 0.0287 \tiny{(-0.0196)} }&\ioub{ 0.0646 \tiny{(-0.0643)} }&\goiou{ 0.0246 \tiny{(-0.0005)} }&\iou{ 0.0592 \tiny{(-0.2781)} }&\ioub{ 0.0008 \tiny{(-0.4995)}} & \ioub{-0.4688}\\

\midrule

\multirow{10}{*}[.4em]{\rotatebox[origin=c]{90}{\footnotesize{COVERAGE}}} 

& \multirow{2}{*}{\rotatebox[origin=c]{90}{\tiny{Original}}} &
   \mcc{0.2747  }&\gomcc{0.0075  }&\gomcc{0.0230 }&\mcc{0.2617 }&\gomcc{0.0062 }&\mcc{0.0615   }&\gomcc{-0.0571  }&\mcc{0.4442}&\gomcc{0.0082} & \mcc{-} \\
& &\iou{0.2199} & \goiou{0.0109} & \goiou{0.0856} & \iou{0.1856} & \goiou{0.0858} & \iou{0.1106} & \goiou{0.0423} & \iou{0.3752} & \goiou{0.0070} & \iou{-} \\[3pt]

& \multirow{2}{*}{\rotatebox[origin=c]{90}{\tiny{CamTE}}} 
  &\mccb{0.1480 \tiny{(-0.1267)}} & \gomcc{0.0056  \tiny{(-0.0020)}} & \gomcc{-0.0015 \tiny{(-0.0245)}} & \mcc{0.0790 \tiny{(-0.1827)}} & \gomcc{-0.0230 \tiny{(-0.0292)}} & \mccb{0.0489 \tiny{(-0.0127)}} & \gomcc{-0.0722 \tiny{(-0.0151)}} & \mccb{0.2614 \tiny{(-0.1828)}} & \gomcc{0.0000 \tiny{(-0.0082)}} & \mccb{-0.1646} \\
& &\ioub{0.1162 \tiny{(-0.1038)}} & \goiou{0.0079 \tiny{(-0.0030)}} & \goiou{0.0711 \tiny{(-0.0145)}} & \iou{0.0719 \tiny{(-0.1137)}} & \goiou{0.0770 \tiny{(-0.0089)}} & \ioub{0.1043 \tiny{(-0.0063)}} & \goiou{0.0361 \tiny{(-0.0062)}} & \ioub{0.2212 \tiny{(-0.1541)}} & \goiou{0.0000 \tiny{(-0.0070)}} & \ioub{-0.1048} \\[5pt]

& \multirow{2}{*}{\rotatebox[origin=c]{90}{\tiny{BM3D}}} 
  &\mcc{0.2666 \tiny{(-0.0081)}} & \gomcc{0.0051  \tiny{(-0.0024)}} & \gomcc{-0.0281 \tiny{(-0.0511)}} & \mccb{0.0371 \tiny{(-0.2246)}} & \gomcc{-0.0145 \tiny{(-0.0207)}} & \mcc{0.0515 \tiny{(-0.0100)}} & \gomcc{-0.0771 \tiny{(-0.0200)}} & \mcc{0.3267 \tiny{(-0.1175)}} & \gomcc{0.0000 \tiny{(-0.0082)}} & \mcc{-0.1141}\\
& &\iou{0.2151 \tiny{(-0.0049)}} & \goiou{0.0036 \tiny{(-0.0072)}} & \goiou{0.0617 \tiny{(-0.0240)}} & \iou{0.0841 \tiny{(-0.1015)}} & \goiou{0.0773 \tiny{(-0.0085)}} & \iou{0.1055 \tiny{(-0.0051)}} & \goiou{0.0336 \tiny{(-0.0087)}} & \iou{0.2863 \tiny{(-0.0889)}} & \goiou{0.0000 \tiny{(-0.0070)}} & \iou{-0.0571} \\[5pt]

& \multirow{2}{*}{\rotatebox[origin=c]{90}{\tiny{Diff-CF}}} 
  &\mccb{0.1598 \tiny{(-0.1149)}} & \gomcc{0.0011  \tiny{(-0.0064)}} & \gomcc{-0.0065 \tiny{(-0.0295)}} & \mccb{0.0483 \tiny{(-0.2133)}} & \gomcc{-0.0115 \tiny{(-0.0176)}} & \mcc{0.0514 \tiny{(-0.0101)}} & \gomcc{-0.0602 \tiny{(-0.0031)}} & \mcc{0.2849 \tiny{(-0.1594)}} & \gomcc{0.0000 \tiny{(-0.0082)}} & \mccb{-0.1595}\\
& &\ioub{0.1278 \tiny{(-0.0922)}} & \goiou{0.0059 \tiny{(-0.0050)}} & \goiou{0.0687 \tiny{(-0.0169)}} & \ioub{0.0537 \tiny{(-0.1320)}} & \goiou{0.0790 \tiny{(-0.0068)}} & \iou{0.1055 \tiny{(-0.0051)}} & \goiou{0.0383 \tiny{(-0.0040)}} & \iou{0.2427 \tiny{(-0.1325)}} & \goiou{0.0000 \tiny{(-0.0070)}} & \ioub{-0.0974} \\[5pt]

& \multirow{2}{*}{\rotatebox[origin=c]{90}{\tiny{Diff-CFG}}} 
  &\mcc{0.2003 \tiny{(-0.0745)}} & \gomcc{-0.0004 \tiny{(-0.0079)}} & \gomcc{-0.0124 \tiny{(-0.0354)}} & \mcc{0.0680 \tiny{(-0.1937)}} & \gomcc{0.0024 \tiny{(-0.0038)} } & \mccb{0.0475 \tiny{(-0.0140)}} & \gomcc{-0.0717 \tiny{(-0.0146)}} & \mccb{0.2738 \tiny{(-0.1704)}} & \gomcc{0.0000 \tiny{(-0.0082)}} & \mcc{-0.1478}\\
& &\iou{0.1607 \tiny{(-0.0592)}} & \goiou{0.0010 \tiny{(-0.0099)}} & \goiou{0.0630 \tiny{(-0.0226)}} & \ioub{0.0693 \tiny{(-0.1163)}} & \goiou{0.0858 \tiny{(0.0000)} } & \ioub{0.1051 \tiny{(-0.0055)}} & \goiou{0.0334 \tiny{(-0.0090)}} & \ioub{0.2386 \tiny{(-0.1367)}} & \goiou{0.0000 \tiny{(-0.0070)}} & \iou{-0.0890} \\

\bottomrule
  \end{tabularx}
  \caption{\iou{IoU} and \mcc{MCC} results for Korus~\cite{korus1, korus2}, FAU~\cite{FAU_dataset} and COVERAGE~\cite{COVERAGE} datasets and all methods, except for Bammey \etal~\cite{bammey}. For each dataset, we present in the first row the performance of the forgery detectors over the original images. Then, in the following rows, we show the performance of the same detectors over the considered counter-forensic versions of the images, and the difference to the original performance ($\text{metric} \,_{CF} - \text{metric}\,_{orig}$). The lower this difference is, the better the counter-forensic method erased the forgery traces. The best two scores are shown in bold and underlined for each database. For the sake of readability, methods that are not able to obtain a reasonable performance over the original dataset (MCC $<$ 0.03) are grayed out. Bammey \etal~\cite{bammey} is excluded from this table, as it was not able to obtain an acceptable performance over any of the considered datasets. The last column (Avg$_w$), is the average of the differences $\text{metric} \,_{CF} - \text{metric}\,_{orig}$, weighted by the performance in the original dataset.\\ 
  }
  \label{tab:all-mcc-iou}
    \vspace*{-0.5 cm}
\end{table*}

For all experiments, we used the following values: $t^*=40$, $s=10^6$, and $\mathcal{D} = $ -- SSIM~\cite{ssim} as the guidance metric. A detailed discussion on the influence of the hyper-parameters is presented in Sec.~\ref{sec:parameters}. In all cases, the images are divided into patches of $256\times 256$ pixels before running the diffusion process. As for the diffusion model, we used a pre-trained class unconditional checkpoint\footnote{\url{https://github.com/openai/guided-diffusion}}. 



\section{Experiments}
\label{sec:experiments}

\begin{figure*}[t]
  \centering
\includegraphics[width = \textwidth]{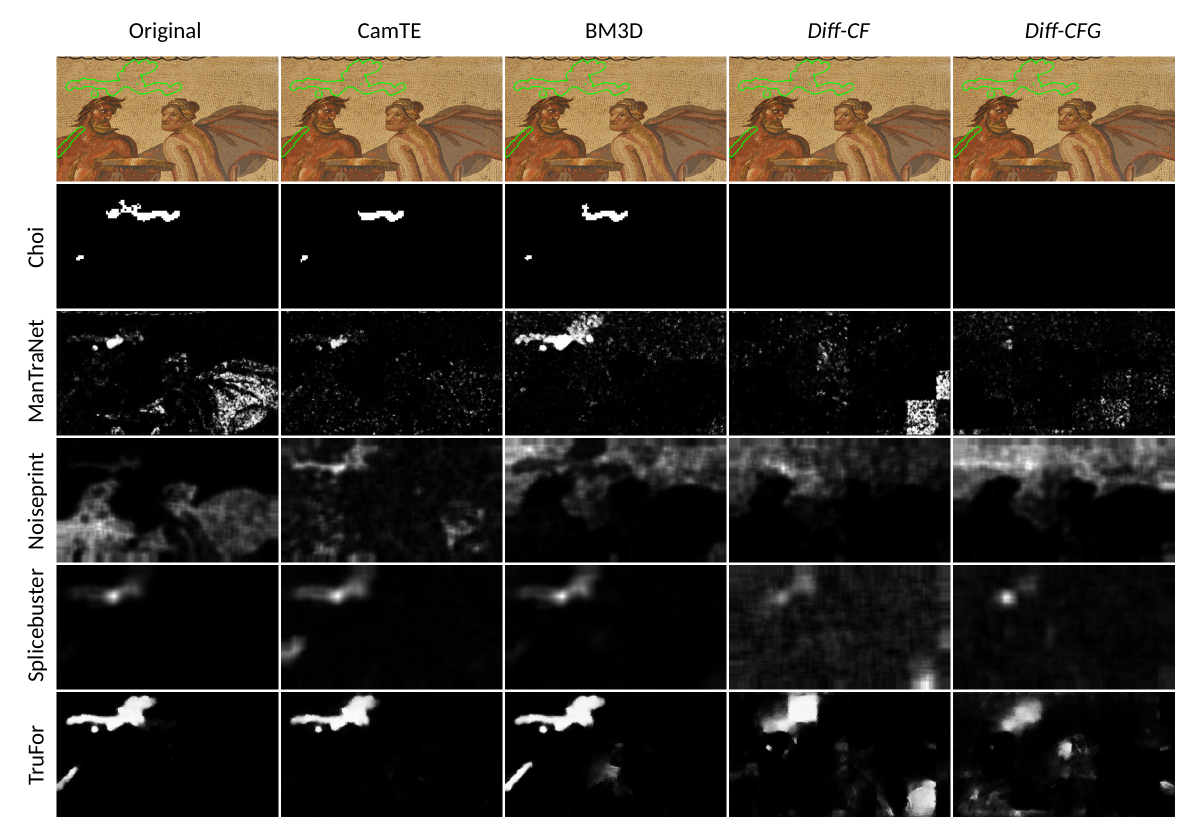}
   \caption{Results obtained by different forensics methods on the different versions of image \texttt{r7710a7fat} from the Korus dataset~\cite{korus1, korus2}. We observe that Choi~\cite{choi}, ManTraNet~\cite{mantranet}, and Noiseprint~\cite{noiseprint} feature no detection when \emph{Diff-CF} or \emph{Diff-CFG} are applied. For Splicebuster~\cite{splicebuster2015} and TruFor~\cite{TruFor}, even if counter-forensics techniques are not completely able to deceive them, the proposed approaches degrade their detections the most. More examples are included in the supplementary materials. }
   \vspace*{-0.4 cm}
   \label{fig:korus_detection}
\end{figure*}

To assess the performance of the proposed approaches, we compared both the non-guided (\emph{Diff-CF}) and the guided (\emph{Diff-CFG}) variants with the Camera Trace Erasing technique (CamTE)~\cite{Chen_2020_CVPR} and with BM3D~\cite{BM3D, ipol.2012.l-bm3d}. While comparison with a plain denoiser is not a common practice in the field, we believe it should be included. Indeed, camera traces are a sort of noise in the sense they produce variations in the pixel's values that are not related to the captured scene. On the other hand, we excluded from the comparison the median filtering, which is a widespread technique in counter-forensics, since it was shown to be outperformed by CamTE~\cite{Chen_2020_CVPR}. 

We ran our comparisons in four image forgery detection benchmark datasets: Korus~\cite{korus1, korus2}, FAU~\cite{FAU_dataset}, COVERAGE~\cite{COVERAGE} and DSO-1~\cite{DSO}. Since most methods except for Bammey \etal~\cite{bammey} deliver poor detection results on the DSO dataset, we decided to exclude them from the main article and report them in the supplementary material.



The goal of counter-forensics methods is to erase all the traces left by the tampering process while preserving the image structures and their semantic content. Therefore, we evaluate two aspects of the counter-forensics techniques under analysis. First, how effectively they hide the forgeries (Sec.~\ref{ref:subsec:tracesremove}) and second, the quality of the purified images (Sec.~\ref{subsec:qualityassesment}).

\subsection{Forgery traces removal}\label{ref:subsec:tracesremove}

The first point to evaluate is how well the proposed approaches remove the forgery traces. To do so, we ran several state-of-the-art forgery detection methods on the original datasets as well as in their counter-forensics versions (images purified using different techniques). To evaluate their capability of deceiving the forensics methods, we look at the difference between the detection performance before and after purification. The forensics methods that were used are: ZERO~\cite{ZERO}, Noiseprint~\cite{noiseprint}, Splicebuster~\cite{splicebuster2015}, ManTraNet~\cite{mantranet}, Choi~\cite{choi, ipol_choi}, Bammey~\cite{bammey}, Shin~\cite{shin2017color}, Comprint~\cite{mareen2022comprint}, CAT-Net~\cite{catnet1, catnet2} and TruFor~\cite{TruFor}. A brief description of each method can be found in the supplementary material. 


To measure detections, we provide scores with the Intersection over Union (IoU) and the Matthews Correlation Coefficient (MCC). F1 scores are not included in the main article but are available in the supplementary material. In terms of true positives (TP), true negatives (TN), false positives (FP), and false negatives (FN), the IoU is the ratio between the number of pixels in the intersection of detected samples and of ground-truth-positive samples and the number of pixels in the union of these sets. On the other hand, the MCC represents the correlation between the ground truth and detections. The definitions of both the IoU and MCC scores are given in the supplementary material.

The scores were computed for each image and then averaged over each dataset. As most surveyed methods do not provide a binary output but a heat map, to adapt the metrics to the continuous setting, we used their weighted version. We regard the value of a heat map $H$ at each pixel $u$ as the probability of forgery of the pixel. Therefore, given the ground truth mask $M$, we define the \textit{weighted} TP, \textit{weighted} FP, \textit{weighted} TN and \textit{weighted} FN as:
\begin{align}
TP_w &= \sum_{u} H(u) \cdot M(u), \\
FP_w &= \sum_x (1-H(u)) \cdot M(u),\\
TN_w &= \sum_x (1-H(u)) \cdot (1-M(u)), \\
FN_w &= \sum_x H(u) \cdot (1-M(u)).
\end{align}

IoU and MCC results for all datasets and all methods are presented in Tab.~\ref{tab:all-mcc-iou}. 
For each dataset, we present in the first row the performance of the forgery detectors over the original images. Then, in the following rows, we show the performance of the same detectors over the considered counter-forensic versions of the images and the difference to the original performance ($\text{metric} \,_{purified} - \text{metric}\,_{orig}$). The lower this difference is, the better the counter-forensic method erased the forgery traces.

The results in Tab.~\ref{tab:all-mcc-iou} show that the proposed counter-forensic methods based on diffusion models outperform other counter-forensic techniques in most cases. Indeed, except for the COVERAGE dataset~\cite{COVERAGE} where our methods rank second and third (after CamTE), in all the rest of the datasets \textit{Diff-CF} and \textit{Diff-CFG} achieve the best score reductions. When comparing \textit{Diff-CF} to \textit{Diff-CFG}, we observe that the non-guided version delivers, in most cases, the best results as a counter-forensic method. This can be explained by the fact that when we do not condition the method, the reverse generative process is able to get closer to the distribution of the clean training data.

Regarding the forensic methods individually, we observe that TruFor~\cite{TruFor} outperforms the rest of the methods in most cases. Furthermore, it is the only method that still performs acceptably after applying counter-forensics attacks, except on the FAU dataset~\cite{FAU_dataset}. Indeed, in this case, once counter-forensic methods are applied, the method delivers highly deteriorated results.

Fig.~\ref{fig:korus_detection} shows an example of the results obtained by different forensics methods on the different versions of the same forged image. We observe that Choi delivers nearly the same result as in the original forgery when CamTE or BM3D are applied. However, it features no detection when \textit{Diff-CF} or \textit{Diff-CFG} are used as counter-forensics techniques. Noiseprint and ManTraNet provide better detections when CamTE or BM3D are applied, respectively. However, no detection is made when using the proposed approaches. On the other hand, none of the counter-forensics methods is able to deceive Splicebuster and Trufor completely. However, we can observe that their results degrade the most when \textit{Diff-CF} and \textit{Diff-CFG} are applied \footnote{An analysis of the robustness of these forensic methods is out of the scope of this work and will be addressed in the future.}.

\subsection{Image Quality Assessment}\label{subsec:qualityassesment}

\begin{table}[t]  
  \small  
  \centering
    \setlength{\tabcolsep}{0.4em}
  \begin{tabularx}{0.47\textwidth}{llcc@{\hskip 5pt}| @{\hskip 9pt} ccc}

\toprule

& & NIQE & BRISQUE & LPIPS & PSNR & SSIM \\ 
& & ($\blacktriangledown$) & ($\blacktriangledown$) & ($\blacktriangledown$) & ($\blacktriangle$) & ($\blacktriangle$) \\
\midrule
\multirow{5}{*}[.4em]{\rotatebox[origin=c]{90}{Korus}} & 
Original & 5.7271 & 13.7602 & 0.0000 & 80.0000 & 1.0000 \\
& CamTE  & 5.5442 & 34.5632 & 0.1684 & \textbf{38.2833} & \textbf{0.9433} \\
& BM3D   & 5.1004 & 38.0418 & 0.0835 & \textbf{43.1409} & \textbf{0.9802} \\
& \textit{Diff-CF}  & \textbf{3.8693} & \textbf{23.1161} & \textbf{0.0733} & 32.9680 & 0.8769 \\
& \textit{Diff-CFG} & \textbf{4.1070} & \textbf{28.3290} & \textbf{0.0771} & 34.3391 & 0.9126 \\

\midrule
\multirow{5}{*}[.4em]{\rotatebox[origin=c]{90}{FAU}} & 
Original & 4.7392 & 20.5726 & 0.0000 & 80.0000 & 1.0000 \\
& CamTE  & 5.8360 & 40.1577 & 0.2098 & \textbf{37.8765} & \textbf{0.9460} \\
& BM3D   & 5.4875 & 42.7470 & 0.1045 & \textbf{41.2625} & \textbf{0.9797} \\
& \textit{Diff-CF}  & \textbf{3.8896} & \textbf{19.8268} & \textbf{0.0985} & 33.0308 & 0.8792 \\
& \textit{Diff-CFG} & \textbf{4.2440} & \textbf{29.9920} & \textbf{0.0952} & 34.4725 & 0.9159 \\

\midrule
\multirow{5}{*}[.4em]{\rotatebox[origin=c]{90}{COVERAGE}} & 
Original & 4.5529 & 19.0256 & 0.0000 & 80.0000 & 1.0000 \\
& CamTE  & 5.4513 & 30.3558 & 0.0631 & \textbf{35.7974} & \textbf{0.9648} \\
& BM3D   & 5.8792 & 35.9560 & \textbf{0.0237} & \textbf{44.1417} & \textbf{0.9888} \\
& \textit{Diff-CF}  & \textbf{4.3343} & \textbf{17.1298} & 0.0281 & 33.4959 & 0.9275 \\
& \textit{Diff-CFG} & \textbf{5.0359} & \textbf{27.8903} & \textbf{0.0276} & 34.6969 & 0.9487 \\

\bottomrule
  \end{tabularx}
  \caption{Image quality assessment results of the evaluated counter-forensics techniques. The $\blacktriangledown$ indicates that the lower the score the better while the $\blacktriangle$ indicates that the higher the score the better. The best two scores are shown in bold for each database. For the no-reference metrics NIQE and BRISQE, the proposed diffusion-based counter-forensics methods achieve the best performance.}
  \label{tab:all-iqa}
  \vspace*{-0.3 cm}
\end{table}

\begin{figure*}[t]
  \centering
\includegraphics[width=\textwidth]{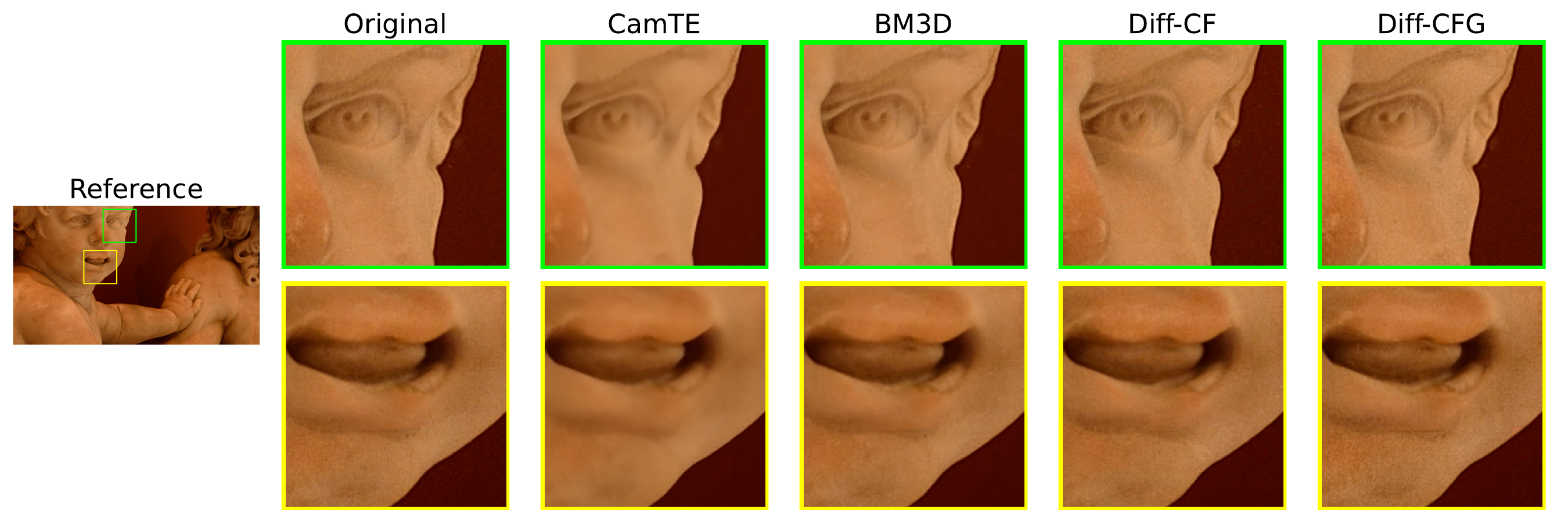}
   \caption{Image quality comparison for all considered counter-forensics methods. We observe that both \textit{Diff-CF} and \textit{Diff-CFG} are good at preserving the fine textures and edges of the image while CamTE and BM3D blur all these fine structures.}
   \label{fig:querubines}
\vspace*{-0.3 cm}
\end{figure*}

Another important point to evaluate the pertinence of counter-forensic methods is their resulting image quality. We evaluate this quality in two senses. Firstly, we are interested in how natural the purified images are. To evaluate this, we use the reference-free image quality assessment techniques NIQE~\cite{NIQE} and BRISQE~\cite{BROSQE}. Secondly, it is also important to measure the similarity between the input image and the one obtained after the counter-forensic attack. We, of course, want these two images to be perceptually similar. To evaluate this aspect, we use the full reference image quality assessment methods LPIPS~\cite{lpips}, SSIM~\cite{ssim}, and PSNR. For all the metrics, we use the implementations provided by the PyIQA library~\cite{pyiqa}. 

Results are presented in Tab.~\ref{tab:all-iqa}. For all reference-free metrics, the proposed diffusion-based counter-forensics methods achieve the best performance. For the full-reference metrics, we also obtained the best performance for LPIPS, but BM3D and CamTE get better performance in terms of PSNR and SSIM. 

Among the proposed methods, the guided variant always achieves better performance in terms of PSNR and SSIM, as expected. Indeed, the guidance explicitly encourages the purified image to be close to the input image. Still, the results are not so conclusive when evaluating the LPIPS score, where the non-guided version shows a slightly better performance on the Korus dataset.

It is important to mention that even if \textit{Diff-CFG} uses SSIM as the guidance distance, this does not imply that the obtained scores on that metric should be perfect. In Eq.~\ref{eq:guided}, the guidance can be interpreted as a sort of gradient descent towards the minimum of $\mathcal{D}(\cdot, x_{in})$. To achieve this minimum, the guidance scale $s_t$ plays a crucial role. Using a non-optimum (in terms of the optimization problem), guidance scale causes the final SSIM score not to be optimal. But this ``optimum'' guidance scale could not be the best to effectively erase the forgery traces. Sec.~\ref{sec:parameters} studied this trade-off more in-depth.

Regarding the reference-free image quality assessment metrics, \textit{Diff-CF} consistently achieves better results than \textit{Diff-CFG}. This can be explained by the fact that the unconstrained generative process gets closer to the distribution of the images with which it was trained. Therefore, these images look more natural.

Fig.~\ref{fig:querubines} shows a qualitative example of the different purified images. We observe that both \textit{Diff-CF} and \textit{Diff-CFG} are good at preserving the fine textures and edges of the image, while CamTE and BM3D blur all these fine structures. For instance, the details highlighted in the green patch show that the granularity in the cherubs' cheeks is blurred out by BM3D and CamTE, while it is preserved by the diffusion-based models. This is also visible in the cherubs' chin, highlighted in the yellow patch. As for the edges, the sharpness of the nose (green patch) and the lips (yellow patch) are also better preserved by the proposed approaches.

\subsection{Influence of the parameters}
\label{sec:parameters}

\begin{figure*}[t]
  \centering
\begin{tabular}{c c c}
   & \footnotesize{Time} & \footnotesize{Guidance scale} \\
 \raisebox{4.7em}{\rotatebox[origin=c]{90}{\footnotesize{Forgery traces removal}}}  & \includegraphics[width= 0.83\columnwidth]{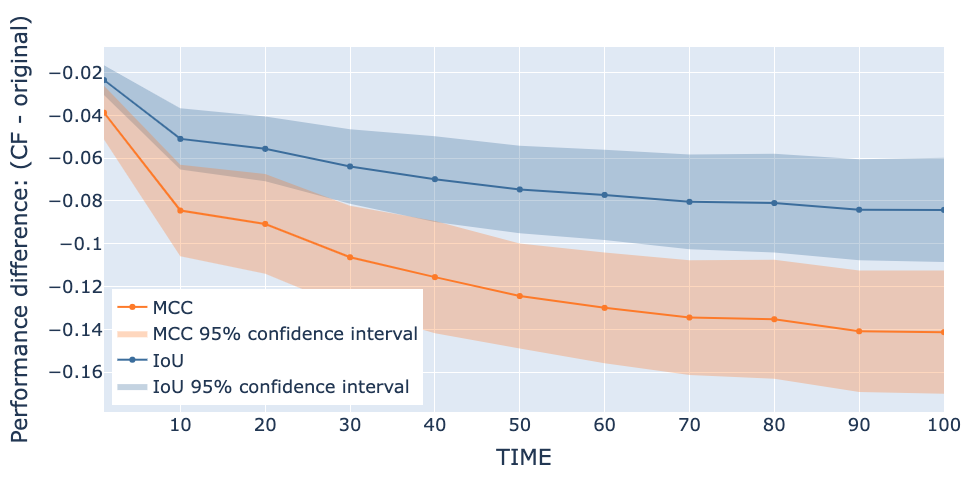} &  \includegraphics[width=0.83\columnwidth]{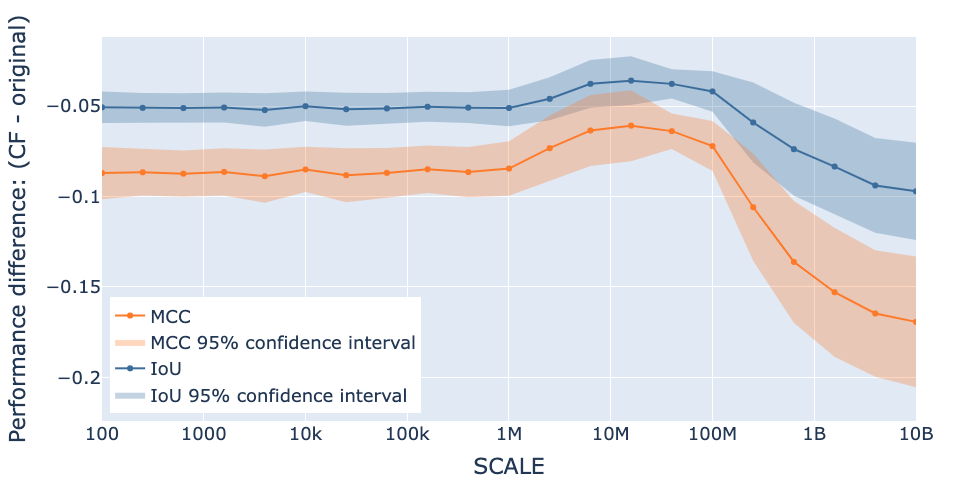}\\
  \raisebox{6em}{\rotatebox[origin=c]{90}{\footnotesize{Image quality}}}  & \includegraphics[width= 0.83\columnwidth]{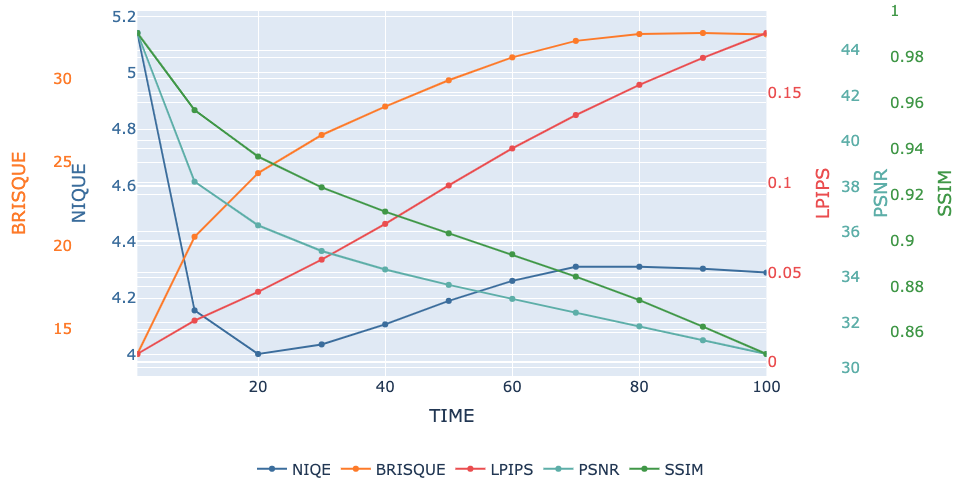} &  \includegraphics[width=0.83\columnwidth]{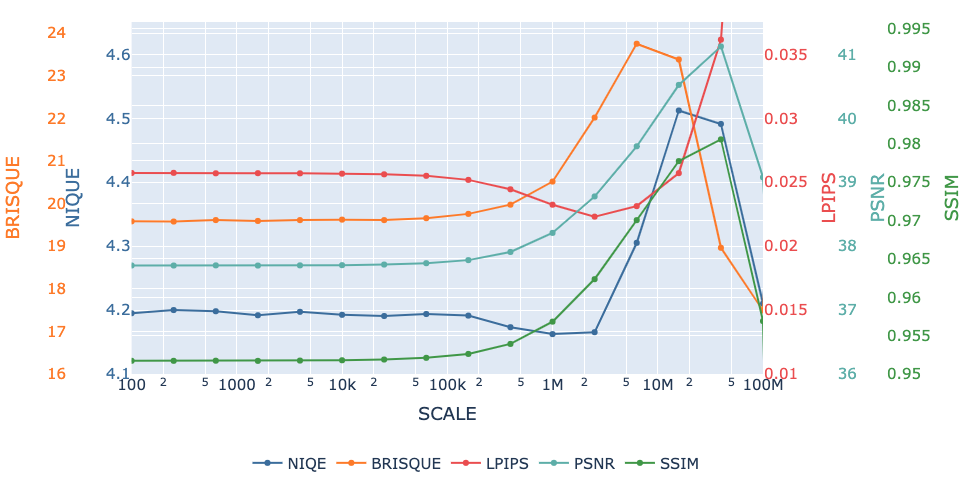}\\
\end{tabular}
 \caption{Study of the impact of the time-step $t^*$ (left-hand side), and guidance scale $s$ (right-hand side). For each parameter, we evaluate its influence on the forgery traces removal task (top) and on the purified image quality (bottom). For the forgery traces removal task, we plot the average difference between the performance before and after purification for the best-performing methods in the original dataset as a function of the parameters' value. The colored background area represents the 95\% confidence interval. For the Image quality assessment, all five metrics presented in Sec.~\ref{subsec:qualityassesment} are plotted as a function of the parameters' value, each with a different axis, for better visualization. This figure is best viewed in color. An interactive version of these plots is included in the supplementary material.}
   \label{fig:time-scale}
\vspace*{-0.3 cm}
 \end{figure*}

The goal of this work is to provide a first study on the use of diffusion models as counter-forensics techniques. As such, it is important to evaluate how the results vary along with the parameters. The non-guided approach \textit{Diff-CF} has only one parameter: the time step $t^*$, while \textit{Diff-CFG} has two: the time step $t^*$ and the guidance scale $s$. In this experiment, we focus mainly on \textit{Diff-CFG} since we think the interaction of both parameters is way more complex than analyzing a single one. The experiments in this section are carried out on Korus dataset~\cite{korus1, korus2}. We evaluate both the forgery traces removal capabilities and the image quality of the purified images. For the first, we compute the performance drop for the best-performing methods over the original dataset: 
Choi, MantraNet, Noiseprint, Splicebuster, and TruFor. For the second, we use all the image quality assessment metrics presented in Sec.~\ref{subsec:qualityassesment}.

\vspace*{-0.3cm}
\paragraph{Diffusion time-step.}
\label{sec:ablation-time}

The results of the impact of the time-step $t^*$ are presented on the left-hand side of Fig.~\ref{fig:time-scale}. The analysis is pretty straightforward: the larger the value of $t^*$, the forgery traces removal performance improves (gets lower). On the other hand, the image quality metrics improve the smaller the value of $t^*$. There is a clear trade-off in the selection of this parameter, that is simple to understand: with higher values of $t^*$, we add more noise to the original image in the forward diffusion process, which makes it easier to hide the forgery traces. On the other hand, starting the reverse process too far away from the original image leads to larger deviations between the original image and the purified one. 

In addition, it is interesting to note that all the full-reference metrics keep strongly degrading as we increase $t^*$, but the reference-free metrics seem to follow a more asymptotic behavior. This evidence can be explained due to the fact that, even if the generated images are more apart from the original one, the diffusion process, following the learned distribution, is still generating natural images. 

\vspace*{-0.3cm}
\paragraph{Guidance scale}

The guidance scale ensures that the purified image remains close to the manipulated image, thus not modifying its semantic content. However, it is crucial that the chosen guidance scale is not excessively large since it would cause the purified image to match the adulterated image, potentially retaining the manipulation traces~\cite{GuidedDiffAdvPurifFROM_RANDOM_NOISE}.

We conducted a series of experiments to study the scale influence, varying the scale value ($s$ in Eq.~\ref{ec:scale}), while keeping a fixed time-step $t^*=10$. As can be seen in the right-hand side of Fig.~\ref{fig:time-scale}, the performance difference has small variations for about the first half of the scale range studied, then shows a slight increase, and finally, a great drop. The best point we could choose would be with the lowest value, so at first, one could be tempted to use the highest value for the scale. But if we add the image quality assessment to the analysis, we observe that for those scale values, the quality of the images is highly degraded. Therefore, an intermediate point should be chosen. Note that the optimal point for the removal of forgery traces is not the optimal point in terms of image quality. As mentioned in Sec.~\ref{subsec:qualityassesment}, this could explain why, in our experiments, we do not obtain the best performance in terms of SSIM, even though we are guiding the diffusion process with this metric.


\section{Conclusions and Future Work}

In this article, we presented a first study on the use of diffusion models for counter-forensics tasks. We showed that such an approach can deliver better results than the existing techniques for both forgery trace removal and image quality. Of course, there is a risk that the shown approaches would be used by people wanting to create forgeries and make them look authentic. The simplicity of this method increases this risk. However, it is also because of its simplicity that the method should be made public: It is important to expose the shortcomings of current methods so that one can know how much trust can be put into an image and so that alternative ways of authentication are developed.

In this direction, future work includes analyzing the traces left by the diffusion purification process~\cite{corvi2022detection} to check whether the use of such a counter-forensic approach can be detected or not. Also, it would be interesting to analyze the robustness of the different methods to such kind of counter-forensic methods.


{\small
\bibliographystyle{ieee_fullname}
\bibliography{egbib}
}

\end{document}


\title{Diffusion models meet image counter-forensics\\
Suplementary material}

\author{Matías Tailanián$^{1,2}$, Marina Gardella$^3$, Alvaro Pardo$^{2,4}$, Pablo Musé$^{1}$\\
$^1$IIE, Facultad de Ingeniería,Universidad de la República\\
$^2$Digital Sense\\
$^3$Centre Borelli, École Normale Supérieure Paris-Saclay, Université Paris-Saclay\\
$^4$Universidad Católica del Uruguay\\
}

\maketitle

\section{Forensic methods}

In this section, we provide a brief description of the forensic methods used to analyze the effectiveness of the counter-forensic approaches. 

Choi et al.~\cite{choi} aims to detect inconsistencies in the mosaic pattern with which the raw image was captured. To do so, they use the fact that sampled pixels were more likely to take extreme values. Also aiming at demosaicing inconsistencies, Shin et al.~\cite{shin2017color} use the fact that sampled pixels have a higher variance to detect forged regions. Bammey et al.~\cite{bammey} combined the translation invariance of convolutional neural networks with the periodicity of the mosaic pattern to train a self-supervised network into implicitly detecting demosaicing artefacts.

Splicebuster~\cite{splicebuster2015} uses the co-occurrences of noise residuals as local features revealing tampered image regions. Noiseprint~\cite{noiseprint} extends on Splicebuster and uses a Siamese network trained on authentic images to extract the noise residual of an image, which is then analyzed for inconsistencies. TruFor~\cite{TruFor} also relies on a noise-sensitive fingerprint that is used with the RGB image to detect deviations from the expected regular pattern that characterizes each pristine image. 

Zero~\cite{ZERO} targets JPEG artifacts. This method counts the number of null DCT coefficients in all blocks and deduces the grid origin. By doing this locally, this method can detect regions having an inconsistent grid origin. Comprint~\cite{mareen2022comprint} combines the use of a compression fingerprint with the noise fingerprint in~\cite{noiseprint}. Comprint is an end-to-end fully convolutional neural network including RGB and DCT
streams, aiming at learning compression artifacts on RGB and DCT domains jointly. 

ManTraNet~\cite{mantranet} is a bipartite end-to-end network, trained to detect image-level manipulations with one part, while the second part is trained on synthetic forgery datasets to detect and localize forgeries in the image.

\section{Forgery detection scores}
 In terms of true positives (TP), true negatives (TN), false positives (FP), and false negatives (FN), the IoU is the ratio between the number of pixels in the intersection of detected samples and of ground-truth-positive samples and the number of pixels in the union of these sets:
\begin{equation}
    IoU = \frac{TP}{TP +FN + FP},
\end{equation}
while the MCC represents the correlation between the ground truth and detections: 
{\small
\begin{equation} 
\begin{split}
&MCC = \\
&\frac{T P \times T N -  F P \times F N}{\sqrt{(T P + F P ) (T P + F N ) (T N + F P ) (T N + F N )}}.
\end{split}
\end{equation}}

\section{Complete IQA scores}
In this Section, we present the Image Quality Assessment for all CF methods, like in Table~2 of the paper, but also include the results over the DSO-1 dataset~\cite{DSO}. For this dataset, we obtained the best scores for NIQE, BRISQUE, and LPIPS.

\begin{table}[t]  
  \small  
  \centering
    \setlength{\tabcolsep}{0.4em}
  \begin{tabularx}{0.47\textwidth}{llcc@{\hskip 4pt}| @{\hskip 6pt} ccc}

\toprule

& & NIQE & BRISQUE & LPIPS & PSNR & SSIM \\ 
& & ($\blacktriangledown$) & ($\blacktriangledown$) & ($\blacktriangledown$) & ($\blacktriangle$) & ($\blacktriangle$) \\
\midrule
\multirow{5}{*}[.4em]{\rotatebox[origin=c]{90}{Korus}} & 
Original & 5.7271 & 13.7602 & 0.0000 & 80.0000 & 1.0000 \\
& CamTE  & 5.5442 & 34.5632 & 0.1684 & \textbf{38.2833} & \textbf{0.9433} \\
& BM3D   & 5.1004 & 38.0418 & 0.0835 & \textbf{43.1409} & \textbf{0.9802} \\
& \textit{Diff-CF}  & \textbf{3.8693} & \textbf{23.1161} & \textbf{0.0733} & 32.9680 & 0.8769 \\
& \textit{Diff-CFG} & \textbf{4.1070} & \textbf{28.3290} & \textbf{0.0771} & 34.3391 & 0.9126 \\

\midrule
\multirow{5}{*}[.4em]{\rotatebox[origin=c]{90}{FAU}} & 
Original & 4.7392 & 20.5726 & 0.0000 & 80.0000 & 1.0000 \\
& CamTE  & 5.8360 & 40.1577 & 0.2098 & \textbf{37.8765} & \textbf{0.9460} \\
& BM3D   & 5.4875 & 42.7470 & 0.1045 & \textbf{41.2625} & \textbf{0.9797} \\
& \textit{Diff-CF}  & \textbf{3.8896} & \textbf{19.8268} & \textbf{0.0985} & 33.0308 & 0.8792 \\
& \textit{Diff-CFG} & \textbf{4.2440} & \textbf{29.9920} & \textbf{0.0952} & 34.4725 & 0.9159 \\

\midrule
\multirow{5}{*}[.4em]{\rotatebox[origin=c]{90}{COVERAGE}} & 
Original & 4.5529 & 19.0256 & 0.0000 & 80.0000 & 1.0000 \\
& CamTE  & 5.4513 & 30.3558 & 0.0631 & \textbf{35.7974} & \textbf{0.9648} \\
& BM3D   & 5.8792 & 35.9560 & \textbf{0.0237} & \textbf{44.1417} & \textbf{0.9888} \\
& \textit{Diff-CF}  & \textbf{4.3343} & \textbf{17.1298} & 0.0281 & 33.4959 & 0.9275 \\
& \textit{Diff-CFG} & \textbf{5.0359} & \textbf{27.8903} & \textbf{0.0276} & 34.6969 & 0.9487 \\

\midrule
\multirow{5}{*}[.4em]{\rotatebox[origin=c]{90}{DSO-1}} & 
Original & 3.9174 & 16.6183 & 0.0000 & 80.0000 & 1.0000 \\
& CamTE & 5.2180 & 40.2867 & 0.2022 & \textbf{38.5459} & \textbf{0.9446} \\
& BM3D & 5.1870 & 39.8485 & 0.1239 & \textbf{41.9057} & \textbf{0.9750}\\
& \textit{Diff-CFG} & \textbf{3.3907} & \textbf{9.2614} & \textbf{0.0950} & 34.1204 & 0.8862\\
& \textit{Diff-CFG} & \textbf{3.6686} & \textbf{19.3601} & \textbf{0.0899} & 35.3473 & 0.9154 \\

\bottomrule
  \end{tabularx}
  \caption{Image quality assessment results of the evaluated counter-forensics techniques. The $\blacktriangledown$ indicates that the lower the score the better while the $\blacktriangle$ indicates that the higher the score the better. The best two scores are shown in bold and underlined for each database. For the no-reference metrics NIQE and BRISQE, the proposed diffusion-based counter-forensics methods achieve the best performance. This table extends Table~2 of the paper, including the results for DSO-1.}
  \label{tab:all-iqa}
  \vspace*{-0.3 cm}
\end{table}

\section{Complete F1, MCC and IoU scores}

The F1 scores obtained by all the tested methods on the Korus dataset are presented in Table~\ref{tab:korus-f1}. Table~\ref{tab:Korus-MCC} and Table~\ref{tab:korus-iou} present the complete MCC and IoU scores (including Bammey~\etal~\cite{bammey}), respectively.

The F1 scores obtained by all the tested methods on the FAU dataset are presented in Table~\ref{tab:fau-f1}. Table~\ref{tab:fau-mcc} and Table~\ref{tab:fau-iou} present the complete MCC and IoU scores (including Bammey~\etal~\cite{bammey}), respectively.

The F1 scores obtained by all the tested methods on the COVERAGE dataset are presented in Table~\ref{tab:coverage-f1}. Table~\ref{tab:coverage-mcc} and Table~\ref{tab:coverage-iou} present the complete MCC and IoU scores (including Bammey~\etal~\cite{bammey}), respectively.

The F1 scores obtained by all the tested methods on the DSO-1 dataset are presented in Table~\ref{tab:dso-f1}. Table~\ref{tab:dso-mcc} and Table~\ref{tab:dso-iou} present the complete MCC and IoU scores (including Bammey~\etal~\cite{bammey}), respectively.

  \setlength{\tabcolsep}{0.7em}

\begin{table*}[h]  
  \small  
  \centering
  \begin{tabularx}{1\textwidth}{c@{\hskip 2pt}YYYYYYYYYY}

\toprule

 & CatNet & Bammey \etal & Choi & Comprint & MantraNet & Noiseprint & Shin & Splicebuster & TruFor & Zero\\ 
\midrule
Original & 0.0657  & 0.0825  & 0.1717  & 0.0817  & 0.1607  & 0.1389  & 0.1035  & 0.1682  & 0.3473  & 0.0043 \\
CamTE & 0.0420 \tiny{(-0.0237)} & 0.0833 \tiny{(0.0009)} & 0.0562 \tiny{(-0.1155)} & 0.0710 \tiny{(-0.0107)} & 0.1147 \tiny{(-0.0460)} & 0.0987 \tiny{(-0.0402)} & 0.1059 \tiny{(0.0024)} & 0.1261 \tiny{(-0.0422)} & 0.2246 \tiny{(-0.1227)} & 0.0000 \tiny{(-0.0043)}\\
BM3D & 0.0937 \tiny{(0.0280)} & 0.0877 \tiny{(0.0053)} & 0.0314 \tiny{(-0.1403)} & 0.0634 \tiny{(-0.0184)} & 0.1278 \tiny{(-0.0329)} & 0.0939 \tiny{(-0.0450)} & 0.0987 \tiny{(-0.0048)} & 0.1274 \tiny{(-0.0409)} & 0.2803 \tiny{(-0.0670)} & 0.0000 \tiny{(-0.0043)}\\
\textit{Diff-CF} & 0.0324 \tiny{(-0.0333)} & 0.0931 \tiny{(0.0106)} & 0.0440 \tiny{(-0.1278)} & 0.0398 \tiny{(-0.0419)} & 0.0769 \tiny{(-0.0838)} & 0.0669 \tiny{(-0.0720)} & 0.0903 \tiny{(-0.0131)} & 0.0720 \tiny{(-0.0963)} & 0.1841 \tiny{(-0.1631)} & 0.0046 \tiny{(0.0003)}\\
\textit{Diff-CFG} & 0.0775 \tiny{(0.0118)} & 0.0912 \tiny{(0.0087)} & 0.0077 \tiny{(-0.1640)} & 0.0520 \tiny{(-0.0298)} & 0.0980 \tiny{(-0.0627)} & 0.0816 \tiny{(-0.0573)} & 0.0910 \tiny{(-0.0125)} & 0.0852 \tiny{(-0.0831)} & 0.2398 \tiny{(-0.1074)} & 0.0019 \tiny{(-0.0024)}\\

\bottomrule
  \end{tabularx}
  \caption{F1 scores obtained for all the tested methods on the Korus dataset~\cite{korus1, korus2}.}
  \label{tab:korus-f1}

  \begin{tabularx}{1\textwidth}{c@{\hskip 2pt}YYYYYYYYYY}
   & & & & & &  & &  &  & \\ 
  & & & & & &  & &  &  & \\ 

\toprule

 & CatNet & Bammey \etal & Choi & Comprint & MantraNet & Noiseprint & Shin & Splicebuster & TruFor & Zero\\ 
\midrule
Original & 0.0790  & -0.1548  & 0.1971  & 0.0534  & 0.1261  & 0.0988  & 0.0221  & 0.1405  & 0.3428  & 0.0050 \\
CamTE & 0.0468 \tiny{(-0.0322)} & -0.0608 \tiny{(0.0940)} & 0.0597 \tiny{(-0.1374)} & 0.0356 \tiny{(-0.0179)} & 0.0646 \tiny{(-0.0614)} & 0.0420 \tiny{(-0.0569)} & 0.0305 \tiny{(0.0084)} & 0.0817 \tiny{(-0.0588)} & 0.1961 \tiny{(-0.1467)} & 0.0000 \tiny{(-0.0050)}\\
BM3D & 0.0997 \tiny{(0.0207)} & -0.0452 \tiny{(0.1095)} & 0.0352 \tiny{(-0.1619)} & 0.0278 \tiny{(-0.0256)} & 0.0652 \tiny{(-0.0609)} & 0.0420 \tiny{(-0.0569)} & 0.0155 \tiny{(-0.0066)} & 0.0860 \tiny{(-0.0545)} & 0.2579 \tiny{(-0.0850)} & 0.0000 \tiny{(-0.0050)}\\
\textit{Diff-CF} & 0.0418 \tiny{(-0.0371)} & -0.0030 \tiny{(0.1518)} & 0.0147 \tiny{(-0.1825)} & 0.0024 \tiny{(-0.0510)} & 0.0255 \tiny{(-0.1005)} & 0.0190 \tiny{(-0.0798)} & 0.0027 \tiny{(-0.0194)} & 0.0350 \tiny{(-0.1055)} & 0.1454 \tiny{(-0.1974)} & 0.0045 \tiny{(-0.0005)}\\
\textit{Diff-CFG} & 0.0852 \tiny{(0.0063)} & -0.0096 \tiny{(0.1452)} & 0.0044 \tiny{(-0.1927)} & 0.0125 \tiny{(-0.0409)} & 0.0442 \tiny{(-0.0818)} & 0.0267 \tiny{(-0.0722)} & 0.0040 \tiny{(-0.0181)} & 0.0456 \tiny{(-0.0950)} & 0.2064 \tiny{(-0.1364)} & 0.0005 \tiny{(-0.0045)}\\

\bottomrule
  \end{tabularx}
  \caption{MCC scores obtained for all the tested methods (including Bammey~\etal) on the Korus dataset~\cite{korus1, korus2}.}
  \label{tab:Korus-MCC}

  \begin{tabularx}{1\textwidth}{c@{\hskip 2pt}YYYYYYYYYY}
 & & & & & &  & &  &  & \\ 
  & & & & & &  & &  &  & \\ 
\toprule

 & CatNet & Bammey \etal & Choi & Comprint & MantraNet & Noiseprint & Shin & Splicebuster & TruFor & Zero\\ 
\midrule
Original & 0.0433  & 0.0446  & 0.1261  & 0.0461  & 0.0982  & 0.0792  & 0.0568  & 0.1012  & 0.2575  & 0.0028 \\
CamTE & 0.0278 \tiny{(-0.0155)} & 0.0452 \tiny{(0.0005)} & 0.0400 \tiny{(-0.0860)} & 0.0389 \tiny{(-0.0072)} & 0.0644 \tiny{(-0.0338)} & 0.0545 \tiny{(-0.0247)} & 0.0578 \tiny{(0.0011)} & 0.0729 \tiny{(-0.0284)} & 0.1489 \tiny{(-0.1086)} & 0.0000 \tiny{(-0.0028)}\\
BM3D & 0.0646 \tiny{(0.0213)} & 0.0478 \tiny{(0.0031)} & 0.0227 \tiny{(-0.1033)} & 0.0346 \tiny{(-0.0115)} & 0.0746 \tiny{(-0.0237)} & 0.0514 \tiny{(-0.0278)} & 0.0540 \tiny{(-0.0028)} & 0.0744 \tiny{(-0.0268)} & 0.1964 \tiny{(-0.0611)} & 0.0000 \tiny{(-0.0028)}\\
\textit{Diff-CF} & 0.0204 \tiny{(-0.0229)} & 0.0505 \tiny{(0.0059)} & 0.0246 \tiny{(-0.1014)} & 0.0215 \tiny{(-0.0246)} & 0.0416 \tiny{(-0.0566)} & 0.0360 \tiny{(-0.0432)} & 0.0487 \tiny{(-0.0081)} & 0.0401 \tiny{(-0.0611)} & 0.1131 \tiny{(-0.1445)} & 0.0027 \tiny{(-0.0001)}\\
\textit{Diff-CFG} & 0.0527 \tiny{(0.0095)} & 0.0495 \tiny{(0.0048)} & 0.0043 \tiny{(-0.1217)} & 0.0281 \tiny{(-0.0180)} & 0.0552 \tiny{(-0.0430)} & 0.0446 \tiny{(-0.0346)} & 0.0491 \tiny{(-0.0076)} & 0.0488 \tiny{(-0.0525)} & 0.1601 \tiny{(-0.0975)} & 0.0011 \tiny{(-0.0017)}\\

\bottomrule
  \end{tabularx}
  \caption{IoU scores obtained for all the tested methods (including Bammey~\etal) on the Korus dataset~\cite{korus1, korus2}.}
  \label{tab:korus-iou}
\end{table*}
  \setlength{\tabcolsep}{0.7em}

\begin{table*}[h]  
  \small  
  \centering

  \begin{tabularx}{1\textwidth}{c@{\hskip 2pt}YYYYYYYYYY}
\toprule
 & CatNet & Bammey \etal & Choi & Comprint & MantraNet & Noiseprint & Shin & Splicebuster & TruFor & Zero\\ 
\midrule
Original & 0.3187  & 0.0827  & 0.3122  & 0.0566  & 0.0630  & 0.0888  & 0.1843  & 0.0471  & 0.4203  & 0.2958 \\
CamTE & 0.0159 \tiny{(-0.3027)} & 0.0909 \tiny{(0.0082)} & 0.1496 \tiny{(-0.1626)} & 0.0376 \tiny{(-0.0190)} & 0.0787 \tiny{(0.0157)} & 0.0806 \tiny{(-0.0082)} & 0.1639 \tiny{(-0.0204)} & 0.0517 \tiny{(0.0045)} & 0.1008 \tiny{(-0.3196)} & 0.1089 \tiny{(-0.1870)}\\
BM3D & 0.0741 \tiny{(-0.2446)} & 0.0911 \tiny{(0.0084)} & 0.0722 \tiny{(-0.2399)} & 0.0237 \tiny{(-0.0328)} & 0.0690 \tiny{(0.0060)} & 0.0606 \tiny{(-0.0282)} & 0.1298 \tiny{(-0.0545)} & 0.0460 \tiny{(-0.0012)} & 0.1287 \tiny{(-0.2917)} & 0.1012 \tiny{(-0.1946)}\\
\textit{Diff-CF} & 0.0099 \tiny{(-0.3088)} & 0.0973 \tiny{(0.0146)} & 0.0781 \tiny{(-0.2341)} & 0.0294 \tiny{(-0.0271)} & 0.0643 \tiny{(0.0013)} & 0.0383 \tiny{(-0.0505)} & 0.1061 \tiny{(-0.0781)} & 0.0238 \tiny{(-0.0233)} & 0.0935 \tiny{(-0.3269)} & 0.0027 \tiny{(-0.2931)}\\
\textit{Diff-CFG} & 0.0301 \tiny{(-0.2886)} & 0.0953 \tiny{(0.0126)} & 0.0318 \tiny{(-0.2804)} & 0.0275 \tiny{(-0.0291)} & 0.0630 \tiny{(-0.0000)} & 0.0544 \tiny{(-0.0344)} & 0.1109 \tiny{(-0.0733)} & 0.0450 \tiny{(-0.0021)} & 0.1051 \tiny{(-0.3153)} & 0.0016 \tiny{(-0.2942)}\\
\bottomrule
  \end{tabularx}
  \caption{F1 scores obtained for all the tested methods on FAU dataset~\cite{FAU_dataset}.}
  \label{tab:fau-f1}

  \begin{tabularx}{1\textwidth}{c@{\hskip 2pt}YYYYYYYYYY}
& & & & & &  & &  &  & \\ 
& & & & & &  & &  &  & \\ 
\toprule
 & CatNet & Bammey \etal & Choi & Comprint & MantraNet & Noiseprint & Shin & Splicebuster & TruFor & Zero\\ 
\midrule
Original & 0.3228  & -0.1696  & 0.3045  & 0.0393  & 0.0203  & 0.0358  & 0.1134  & 0.0074  & 0.4039  & 0.2757 \\
CamTE & 0.0141 \tiny{(-0.3087)} & -0.0738 \tiny{(0.0958)} & 0.1426 \tiny{(-0.1620)} & 0.0092 \tiny{(-0.0302)} & 0.0154 \tiny{(-0.0049)} & 0.0242 \tiny{(-0.0116)} & 0.0826 \tiny{(-0.0308)} & 0.0045 \tiny{(-0.0029)} & 0.0553 \tiny{(-0.3486)} & -0.0010 \tiny{(-0.2767)}\\
BM3D & 0.0757 \tiny{(-0.2471)} & -0.0692 \tiny{(0.1004)} & 0.0679 \tiny{(-0.2367)} & -0.0017 \tiny{(-0.0410)} & -0.0268 \tiny{(-0.0470)} & -0.0014 \tiny{(-0.0372)} & 0.0411 \tiny{(-0.0723)} & 0.0011 \tiny{(-0.0064)} & 0.0802 \tiny{(-0.3237)} & -0.0114 \tiny{(-0.2871)}\\
\textit{Diff-CF} & 0.0070 \tiny{(-0.3157)} & -0.0130 \tiny{(0.1567)} & 0.0242 \tiny{(-0.2803)} & 0.0001 \tiny{(-0.0392)} & 0.0057 \tiny{(-0.0146)} & -0.0018 \tiny{(-0.0376)} & 0.0128 \tiny{(-0.1006)} & -0.0050 \tiny{(-0.0124)} & 0.0399 \tiny{(-0.3640)} & -0.0007 \tiny{(-0.2765)}\\
\textit{Diff-CFG} & 0.0241 \tiny{(-0.2986)} & -0.0243 \tiny{(0.1453)} & 0.0137 \tiny{(-0.2908)} & -0.0059 \tiny{(-0.0452)} & 0.0128 \tiny{(-0.0075)} & 0.0002 \tiny{(-0.0355)} & 0.0202 \tiny{(-0.0933)} & 0.0127 \tiny{(0.0053)} & 0.0470 \tiny{(-0.3569)} & -0.0043 \tiny{(-0.2800)}\\
\bottomrule
  \end{tabularx}
  \caption{MCC scores obtained for all the tested methods (including Bammey \etal) on FAU dataset~\cite{FAU_dataset}.}
  \label{tab:fau-mcc}

  \begin{tabularx}{1\textwidth}{c@{\hskip 2pt}YYYYYYYYYY}
& & & & & &  & &  &  & \\ 
& & & & & &  & &  &  & \\ 
\toprule
 & CatNet & Bammey \etal & Choi & Comprint & MantraNet & Noiseprint & Shin & Splicebuster & TruFor & Zero\\ 
\midrule
Original & 0.2329  & 0.0469  & 0.2670  & 0.0305  & 0.0336  & 0.0482  & 0.1289  & 0.0251  & 0.3373  & 0.2021 \\
CamTE & 0.0085 \tiny{(-0.2244)} & 0.0516 \tiny{(0.0047)} & 0.1173 \tiny{(-0.1497)} & 0.0206 \tiny{(-0.0099)} & 0.0427 \tiny{(0.0091)} & 0.0434 \tiny{(-0.0049)} & 0.1046 \tiny{(-0.0243)} & 0.0277 \tiny{(0.0026)} & 0.0572 \tiny{(-0.2801)} & 0.0633 \tiny{(-0.1388)}\\
BM3D & 0.0517 \tiny{(-0.1812)} & 0.0517 \tiny{(0.0048)} & 0.0559 \tiny{(-0.2111)} & 0.0126 \tiny{(-0.0179)} & 0.0377 \tiny{(0.0041)} & 0.0331 \tiny{(-0.0152)} & 0.0803 \tiny{(-0.0486)} & 0.0243 \tiny{(-0.0008)} & 0.0799 \tiny{(-0.2574)} & 0.0583 \tiny{(-0.1438)}\\
\textit{Diff-CF} & 0.0056 \tiny{(-0.2273)} & 0.0548 \tiny{(0.0079)} & 0.0458 \tiny{(-0.2213)} & 0.0159 \tiny{(-0.0146)} & 0.0355 \tiny{(0.0019)} & 0.0199 \tiny{(-0.0283)} & 0.0602 \tiny{(-0.0687)} & 0.0123 \tiny{(-0.0128)} & 0.0520 \tiny{(-0.2853)} & 0.0015 \tiny{(-0.2006)}\\
\textit{Diff-CFG} & 0.0184 \tiny{(-0.2145)} & 0.0536 \tiny{(0.0067)} & 0.0220 \tiny{(-0.2451)} & 0.0143 \tiny{(-0.0162)} & 0.0339 \tiny{(0.0003)} & 0.0287 \tiny{(-0.0196)} & 0.0646 \tiny{(-0.0643)} & 0.0246 \tiny{(-0.0005)} & 0.0592 \tiny{(-0.2781)} & 0.0008 \tiny{(-0.2012)}\\

\bottomrule
  \end{tabularx}
  \caption{IoU scores obtained for all the tested methods (including Bammey \etal) on FAU dataset~\cite{FAU_dataset}.}
  \label{tab:fau-iou}
\end{table*}
  \setlength{\tabcolsep}{0.7em}

\begin{table*}[h]  
  \small  
  \centering

\begin{tabularx}{1\textwidth}{c@{\hskip 2pt}YYYYYYYYYY}
\toprule
 & CatNet & Bammey \etal & Choi & Comprint & MantraNet & Noiseprint & Shin & Splicebuster & TruFor & Zero\\ 
\midrule
Original & 0.2971  & 0.1585  & 0.0184  & 0.1508  & 0.2948  & 0.1512  & 0.1948  & 0.0759  & 0.4864  & 0.1890 \\
CameraTE & 0.1651 \tiny{(-0.1320)} & 0.1593 \tiny{(0.0008)} & 0.0128 \tiny{(-0.0056)} & 0.1263 \tiny{(-0.0245)} & 0.1287 \tiny{(-0.1661)} & 0.1360 \tiny{(-0.0152)} & 0.1846 \tiny{(-0.0102)} & 0.0661 \tiny{(-0.0098)} & 0.3220 \tiny{(-0.1644)} & 0.1830 \tiny{(-0.0060)}\\
BM3D & 0.2934 \tiny{(-0.0037)} & 0.1602 \tiny{(0.0018)} & 0.0065 \tiny{(-0.0120)} & 0.1118 \tiny{(-0.0390)} & 0.1485 \tiny{(-0.1463)} & 0.1379 \tiny{(-0.0133)} & 0.1870 \tiny{(-0.0079)} & 0.0613 \tiny{(-0.0145)} & 0.3879 \tiny{(-0.0985)} & 0.1830 \tiny{(-0.0060)}\\
\textit{Diff-CF} & 0.1797 \tiny{(-0.1173)} & 0.1652 \tiny{(0.0067)} & 0.0103 \tiny{(-0.0082)} & 0.1222 \tiny{(-0.0286)} & 0.0961 \tiny{(-0.1987)} & 0.1404 \tiny{(-0.0109)} & 0.1867 \tiny{(-0.0082)} & 0.0692 \tiny{(-0.0067)} & 0.3377 \tiny{(-0.1488)} & 0.0000 \tiny{(-0.1890)}\\
\textit{Diff-CFG} & 0.2253 \tiny{(-0.0717)} & 0.1571 \tiny{(-0.0014)} & 0.0019 \tiny{(-0.0166)} & 0.1136 \tiny{(-0.0373)} & 0.1248 \tiny{(-0.1700)} & 0.1510 \tiny{(-0.0002)} & 0.1859 \tiny{(-0.0090)} & 0.0609 \tiny{(-0.0149)} & 0.3374 \tiny{(-0.1490)} & 0.0000 \tiny{(-0.1890)}\\
\bottomrule
  \end{tabularx}
  \caption{F1 scores for all tested methods on the COVERAGE dataset~\cite{COVERAGE}.}
  \label{tab:coverage-f1}

\begin{tabularx}{1\textwidth}{c@{\hskip 2pt}YYYYYYYYYY}
& & & & & &  & &  &  & \\ 
& & & & & &  & &  &  & \\ 
\toprule
 & CatNet & Bammey \etal & Choi & Comprint & MantraNet & Noiseprint & Shin & Splicebuster & TruFor & Zero\\ 
\midrule
Original & 0.2747  & -0.0376  & 0.0075  & 0.0230  & 0.2617  & 0.0062  & 0.0615  & -0.0571  & 0.4442  & 0.0326 \\
CameraTE & 0.1480 \tiny{(-0.1267)} & -0.0329 \tiny{(0.0047)} & 0.0056 \tiny{(-0.0020)} & -0.0015 \tiny{(-0.0245)} & 0.0790 \tiny{(-0.1827)} & -0.0230 \tiny{(-0.0292)} & 0.0489 \tiny{(-0.0127)} & -0.0722 \tiny{(-0.0151)} & 0.2614 \tiny{(-0.1828)} & 0.0238 \tiny{(-0.0088)}\\
BM3D & 0.2666 \tiny{(-0.0081)} & -0.0415 \tiny{(-0.0039)} & 0.0051 \tiny{(-0.0024)} & -0.0281 \tiny{(-0.0511)} & 0.0371 \tiny{(-0.2246)} & -0.0145 \tiny{(-0.0207)} & 0.0515 \tiny{(-0.0100)} & -0.0771 \tiny{(-0.0200)} & 0.3267 \tiny{(-0.1175)} & 0.0236 \tiny{(-0.0091)}\\
\textit{Diff-CF} & 0.1598 \tiny{(-0.1149)} & -0.0238 \tiny{(0.0138)} & 0.0011 \tiny{(-0.0064)} & -0.0065 \tiny{(-0.0295)} & 0.0483 \tiny{(-0.2133)} & -0.0115 \tiny{(-0.0176)} & 0.0514 \tiny{(-0.0101)} & -0.0602 \tiny{(-0.0031)} & 0.2849 \tiny{(-0.1594)} & 0.0000 \tiny{(-0.0326)}\\
\textit{Diff-CFG} & 0.2003 \tiny{(-0.0745)} & -0.0405 \tiny{(-0.0029)} & -0.0004 \tiny{(-0.0079)} & -0.0124 \tiny{(-0.0354)} & 0.0680 \tiny{(-0.1937)} & 0.0024 \tiny{(-0.0038)} & 0.0475 \tiny{(-0.0140)} & -0.0717 \tiny{(-0.0146)} & 0.2738 \tiny{(-0.1704)} & 0.0000 \tiny{(-0.0326)}\\
\bottomrule
  \end{tabularx}
  \caption{MCC scores for all tested methods (including Bammey \etal) on the COVERAGE dataset~\cite{COVERAGE}.}
  \label{tab:coverage-mcc}

\begin{tabularx}{1\textwidth}{c@{\hskip 2pt}YYYYYYYYYY}
& & & & & &  & &  &  & \\ 
& & & & & &  & &  &  & \\ 
\toprule
 & CatNet & Bammey \etal & Choi & Comprint & MantraNet & Noiseprint & Shin & Splicebuster & TruFor & Zero\\ 
\midrule
Original & 0.2199  & 0.0879  & 0.0109  & 0.0856  & 0.1856  & 0.0858  & 0.1106  & 0.0423  & 0.3752  & 0.1082 \\
CamTE & 0.1162 \tiny{(-0.1038)} & 0.0883 \tiny{(0.0003)} & 0.0079 \tiny{(-0.0030)} & 0.0711 \tiny{(-0.0145)} & 0.0719 \tiny{(-0.1137)} & 0.0770 \tiny{(-0.0089)} & 0.1043 \tiny{(-0.0063)} & 0.0361 \tiny{(-0.0062)} & 0.2212 \tiny{(-0.1541)} & 0.1046 \tiny{(-0.0036)}\\
BM3D & 0.2151 \tiny{(-0.0049)} & 0.0891 \tiny{(0.0012)} & 0.0036 \tiny{(-0.0072)} & 0.0617 \tiny{(-0.0240)} & 0.0841 \tiny{(-0.1015)} & 0.0773 \tiny{(-0.0085)} & 0.1055 \tiny{(-0.0051)} & 0.0336 \tiny{(-0.0087)} & 0.2863 \tiny{(-0.0889)} & 0.1046 \tiny{(-0.0036)}\\
\textit{Diff-CF} & 0.1278 \tiny{(-0.0922)} & 0.0921 \tiny{(0.0041)} & 0.0059 \tiny{(-0.0050)} & 0.0687 \tiny{(-0.0169)} & 0.0537 \tiny{(-0.1320)} & 0.0790 \tiny{(-0.0068)} & 0.1055 \tiny{(-0.0051)} & 0.0383 \tiny{(-0.0040)} & 0.2427 \tiny{(-0.1325)} & 0.0000 \tiny{(-0.1082)}\\
\textit{Diff-CFG} & 0.1607 \tiny{(-0.0592)} & 0.0871 \tiny{(-0.0009)} & 0.0010 \tiny{(-0.0099)} & 0.0630 \tiny{(-0.0226)} & 0.0693 \tiny{(-0.1163)} & 0.0858 \tiny{(0.0000)} & 0.1051 \tiny{(-0.0055)} & 0.0334 \tiny{(-0.0090)} & 0.2386 \tiny{(-0.1367)} & 0.0000 \tiny{(-0.1082)}\\
\bottomrule
  \end{tabularx}
  \caption{IoU scores for all tested methods (including Bammey \etal) on the COVERAGE dataset~\cite{COVERAGE}.}
  \label{tab:coverage-iou}
  
\end{table*}
  \setlength{\tabcolsep}{0.7em}

\begin{table*}[h]  
  \small  
  \centering

\begin{tabularx}{1\textwidth}{c@{\hskip 2pt}YYYYYYYYYY}
\toprule

& CatNet & Bammey \etal & Choi & Comprint & MantraNet & Noiseprint & Shin & Splicebuster & TruFor & Zero\\ 

\midrule
Original & 0.0300  & 0.6876  & 0.0638  & 0.0341  & 0.0853  & 0.0802  & 0.4881  & 0.0600  & 0.0655  & 0.0028 \\
CameraTE & 0.0069 \tiny{(-0.0231)} & 0.6604 \tiny{(-0.0272)} & 0.0490 \tiny{(-0.0148)} & 0.0522 \tiny{(0.0181)} & 0.2487 \tiny{(0.1634)} & 0.1317 \tiny{(0.0515)} & 0.5070 \tiny{(0.0189)} & 0.0868 \tiny{(0.0269)} & 0.2536 \tiny{(0.1881)} & 0.0009 \tiny{(-0.0018)}\\
BM3D & 0.0114 \tiny{(-0.0186)} & 0.6264 \tiny{(-0.0612)} & 0.0156 \tiny{(-0.0482)} & 0.0575 \tiny{(0.0234)} & 0.4571 \tiny{(0.3718)} & 0.1312 \tiny{(0.0509)} & 0.5161 \tiny{(0.0280)} & 0.0909 \tiny{(0.0309)} & 0.3338 \tiny{(0.2683)} & 0.0020 \tiny{(-0.0007)}\\
\textit{Diff-CF} & 0.0035 \tiny{(-0.0265)} & 0.5874 \tiny{(-0.1002)} & 0.3084 \tiny{(0.2446)} & 0.0500 \tiny{(0.0159)} & 0.1431 \tiny{(0.0577)} & 0.0898 \tiny{(0.0095)} & 0.5182 \tiny{(0.0301)} & 0.0604 \tiny{(0.0004)} & 0.3661 \tiny{(0.3006)} & 0.0025 \tiny{(-0.0003)}\\
\textit{Diff-CFG} & 0.0081 \tiny{(-0.0219)} & 0.5908 \tiny{(-0.0969)} & 0.1298 \tiny{(0.0660)} & 0.0652 \tiny{(0.0312)} & 0.1785 \tiny{(0.0931)} & 0.1177 \tiny{(0.0375)} & 0.5180 \tiny{(0.0299)} & 0.0681 \tiny{(0.0081)} & 0.4155 \tiny{(0.3500)} & 0.0043 \tiny{(0.0015)}\\

\bottomrule
  \end{tabularx}
  \caption{F1 scores for all tested methods on the DSO-1 dataset~\cite{DSO}.}
  \label{tab:dso-f1}

\begin{tabularx}{1\textwidth}{c@{\hskip 2pt}YYYYYYYYYY}
& & & & & &  & &  &  & \\ 
& & & & & &  & &  &  & \\ 
\toprule

& CatNet & Bammey \etal & Choi & Comprint & MantraNet & Noiseprint & Shin & Splicebuster & TruFor & Zero\\ 

\midrule
Original & -0.4471  & 0.0796  & -0.0118  & -0.3016  & -0.3010  & -0.3361  & -0.0187  & -0.2838  & -0.8280  & -0.1081 \\
CameraTE & -0.0035 \tiny{(0.4435)} & 0.0577 \tiny{(-0.0219)} & 0.0062 \tiny{(0.0180)} & -0.0682 \tiny{(0.2334)} & -0.1165 \tiny{(0.1845)} & -0.0386 \tiny{(0.2975)} & -0.0037 \tiny{(0.0151)} & -0.1091 \tiny{(0.1746)} & -0.3528 \tiny{(0.4752)} & 0.0016 \tiny{(0.1098)}\\
BM3D & -0.0425 \tiny{(0.4046)} & 0.0070 \tiny{(-0.0726)} & -0.0041 \tiny{(0.0077)} & -0.0251 \tiny{(0.2765)} & -0.0795 \tiny{(0.2215)} & -0.0204 \tiny{(0.3157)} & -0.0035 \tiny{(0.0152)} & -0.1109 \tiny{(0.1729)} & -0.3161 \tiny{(0.5119)} & -0.0049 \tiny{(0.1033)}\\
\textit{Diff-CF} & -0.0018 \tiny{(0.4452)} & 0.0014 \tiny{(-0.0782)} & 0.0228 \tiny{(0.0346)} & -0.0331 \tiny{(0.2685)} & -0.0621 \tiny{(0.2389)} & -0.0498 \tiny{(0.2864)} & -0.0026 \tiny{(0.0162)} & -0.0990 \tiny{(0.1847)} & -0.2850 \tiny{(0.5430)} & -0.0019 \tiny{(0.1063)}\\
\textit{Diff-CFG} & -0.0186 \tiny{(0.4285)} & 0.0010 \tiny{(-0.0785)} & 0.0109 \tiny{(0.0227)} & -0.0473 \tiny{(0.2543)} & -0.1584 \tiny{(0.1426)} & -0.0619 \tiny{(0.2742)} & 0.0028 \tiny{(0.0215)} & -0.1182 \tiny{(0.1656)} & -0.3032 \tiny{(0.5248)} & -0.0003 \tiny{(0.1078)}\\

\bottomrule
  \end{tabularx}
  \caption{MCC scores for all tested methods (including Bammey \etal) on the DSO-1 dataset~\cite{DSO}.}
  \label{tab:dso-mcc}

\begin{tabularx}{1\textwidth}{c@{\hskip 2pt}YYYYYYYYYY}
& & & & & &  & &  &  & \\ 
& & & & & &  & &  &  & \\ 
\toprule

& CatNet & Bammey \etal & Choi & Comprint & MantraNet & Noiseprint & Shin & Splicebuster & TruFor & Zero\\ 
\midrule
Original & 0.0155  & 0.5260  & 0.0396  & 0.0186  & 0.0468  & 0.0438  & 0.3238  & 0.0316  & 0.0358  & 0.0015 \\
CameraTE & 0.0035 \tiny{(-0.0120)} & 0.4943 \tiny{(-0.0317)} & 0.0310 \tiny{(-0.0086)} & 0.0280 \tiny{(0.0094)} & 0.1457 \tiny{(0.0990)} & 0.0724 \tiny{(0.0286)} & 0.3402 \tiny{(0.0164)} & 0.0464 \tiny{(0.0147)} & 0.1512 \tiny{(0.1154)} & 0.0005 \tiny{(-0.0011)}\\
BM3D & 0.0058 \tiny{(-0.0097)} & 0.4570 \tiny{(-0.0690)} & 0.0090 \tiny{(-0.0307)} & 0.0308 \tiny{(0.0123)} & 0.3045 \tiny{(0.2577)} & 0.0721 \tiny{(0.0283)} & 0.3484 \tiny{(0.0245)} & 0.0491 \tiny{(0.0175)} & 0.2145 \tiny{(0.1787)} & 0.0010 \tiny{(-0.0005)}\\
\textit{Diff-CF} & 0.0018 \tiny{(-0.0137)} & 0.4162 \tiny{(-0.1098)} & 0.2025 \tiny{(0.1629)} & 0.0265 \tiny{(0.0079)} & 0.0788 \tiny{(0.0320)} & 0.0479 \tiny{(0.0041)} & 0.3502 \tiny{(0.0263)} & 0.0317 \tiny{(0.0000)} & 0.2354 \tiny{(0.1996)} & 0.0013 \tiny{(-0.0003)}\\
\textit{Diff-CFG} & 0.0041 \tiny{(-0.0114)} & 0.4198 \tiny{(-0.1062)} & 0.0843 \tiny{(0.0446)} & 0.0348 \tiny{(0.0162)} & 0.1048 \tiny{(0.0580)} & 0.0643 \tiny{(0.0205)} & 0.3501 \tiny{(0.0263)} & 0.0357 \tiny{(0.0041)} & 0.2794 \tiny{(0.2436)} & 0.0022 \tiny{(0.0007)}\\

\bottomrule
  \end{tabularx}
  \caption{IoU scores for all tested methods (including Bammey \etal) on the DSO-1 dataset~\cite{DSO}.}
  \label{tab:dso-iou}
  
\end{table*}

\section{Interactive plots}

For better visualization, all plots of Section~5.3 of the paper, are included in separate \verb|.html| files, in interactive versions, as part of the supplementary materials.

\section{Visual results}
\begin{figure*}[t]
  \centering
\includegraphics[width = .95\textwidth]{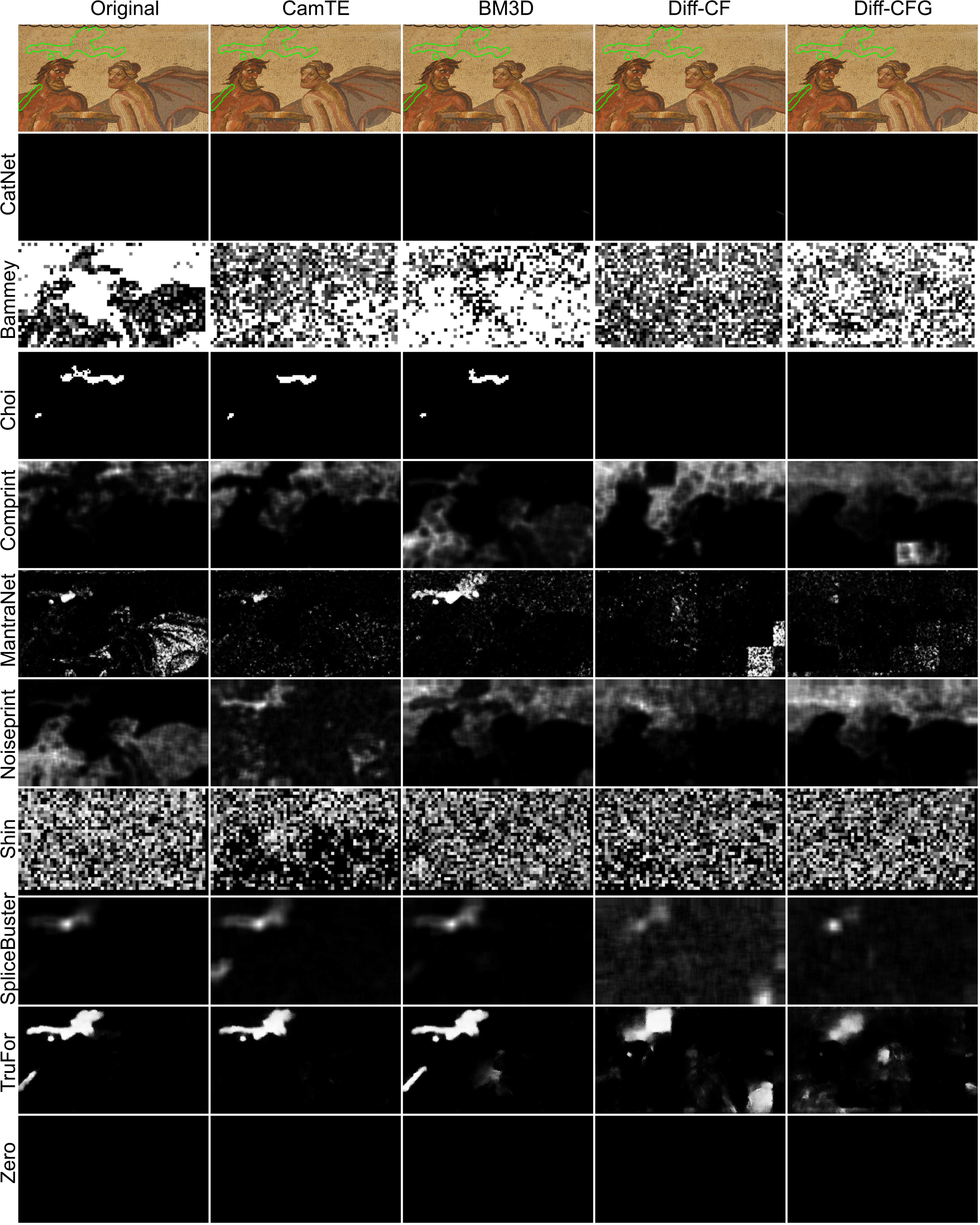}
   \caption{Results obtained by all considered forensics methods on the different versions of image \texttt{r7710a7fat} from the Korus dataset~\cite{korus1, korus2}. This is the same image shown in Figure~1 in the paper, but with the results for all methods, even if they do not detect anything in the original image.}
   \label{fig:r7710a7fat}
\end{figure*}

\begin{figure*}[t]
  \centering
\includegraphics[width = .9\textwidth]{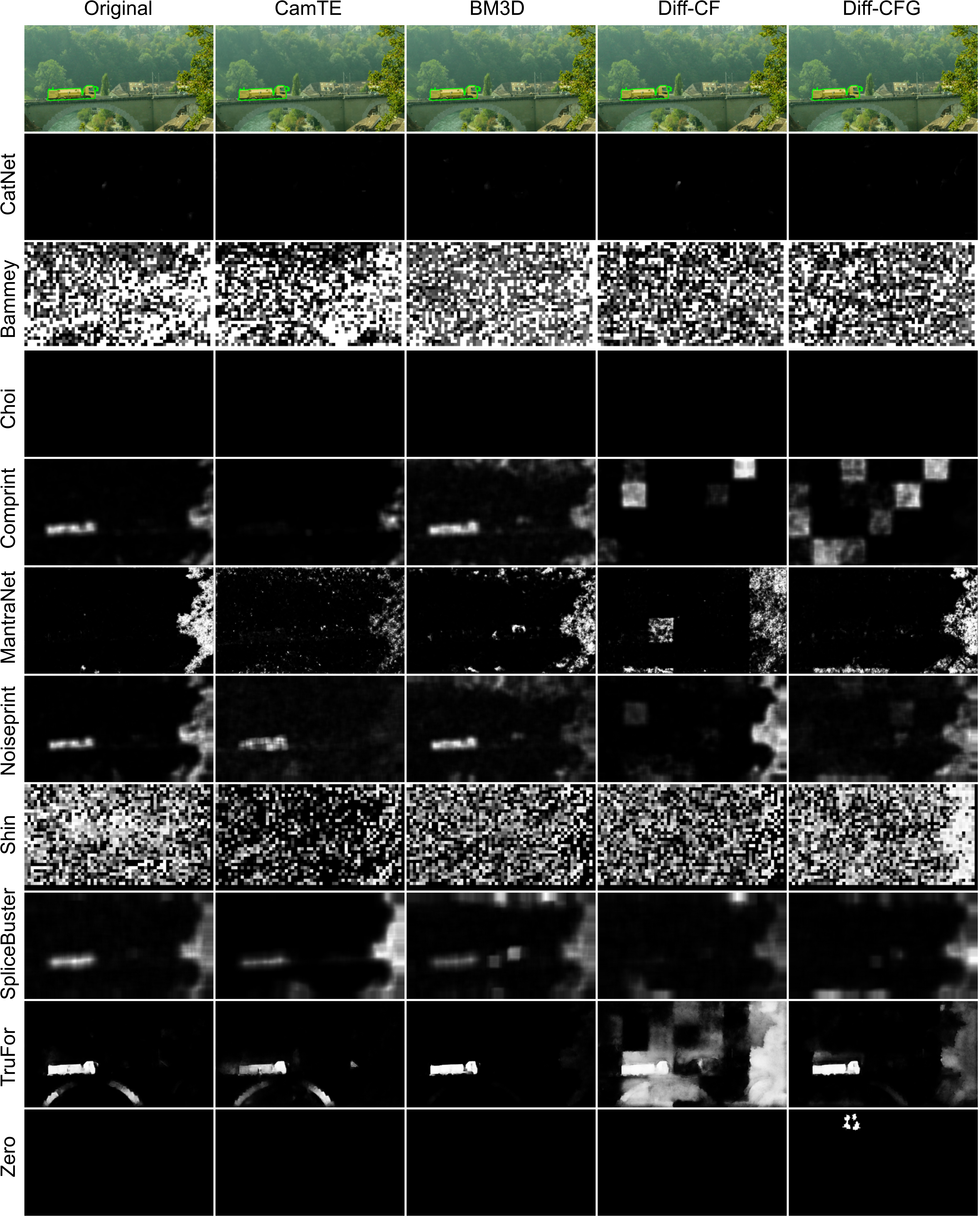}
   \caption{Results obtained by different forensics methods on the different versions of image \texttt{r02354285t} from the Korus dataset~\cite{korus1, korus2}. In this image we observe that Noiseprintt, Splicebuster and TruFor give fairly correct detections in the original forged image. Splicebuster and Noiseprint present degraded detections once BM3D or CamTE are applied. However, the forgery is not even highlighted when \emph{Diff-CF} or \emph{Diff-CFG} are used as counter-forensics attacks. TruFor is more robust to such attacks. Still, their results degrade after \emph{Diff-CF}}
   \label{fig:r02354285t}
\end{figure*}

\begin{figure*}[t]
  \centering
\includegraphics[width = .8\textwidth]{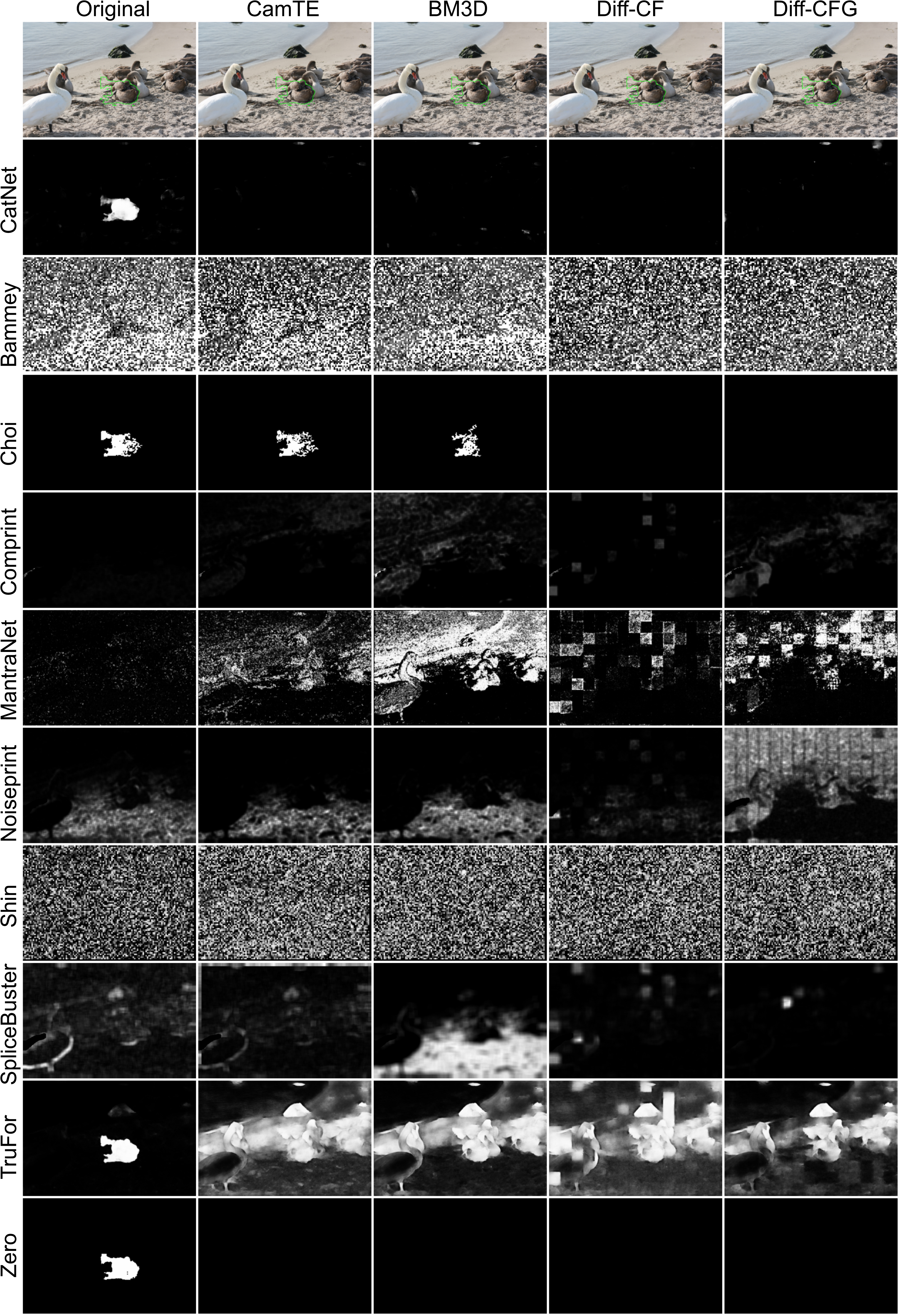}
   \caption{Results obtained by different forensics methods on the different versions of image \texttt{swan\_copy} from the FAU dataset~\cite{FAU_dataset}.}
   \label{fig:swan}
\end{figure*}

Figures~\ref{fig:r7710a7fat}-\ref{fig:swan} present supplementary examples of the results obtained by the different forensics methods on the different versions of the very same image. The image used in each figure is specified in the caption. For all figures, we present the result of all considered forensic methods, even if they do not perform a good detection.

\begin{figure*}[t]
  \centering
\includegraphics[width = .9\textwidth]{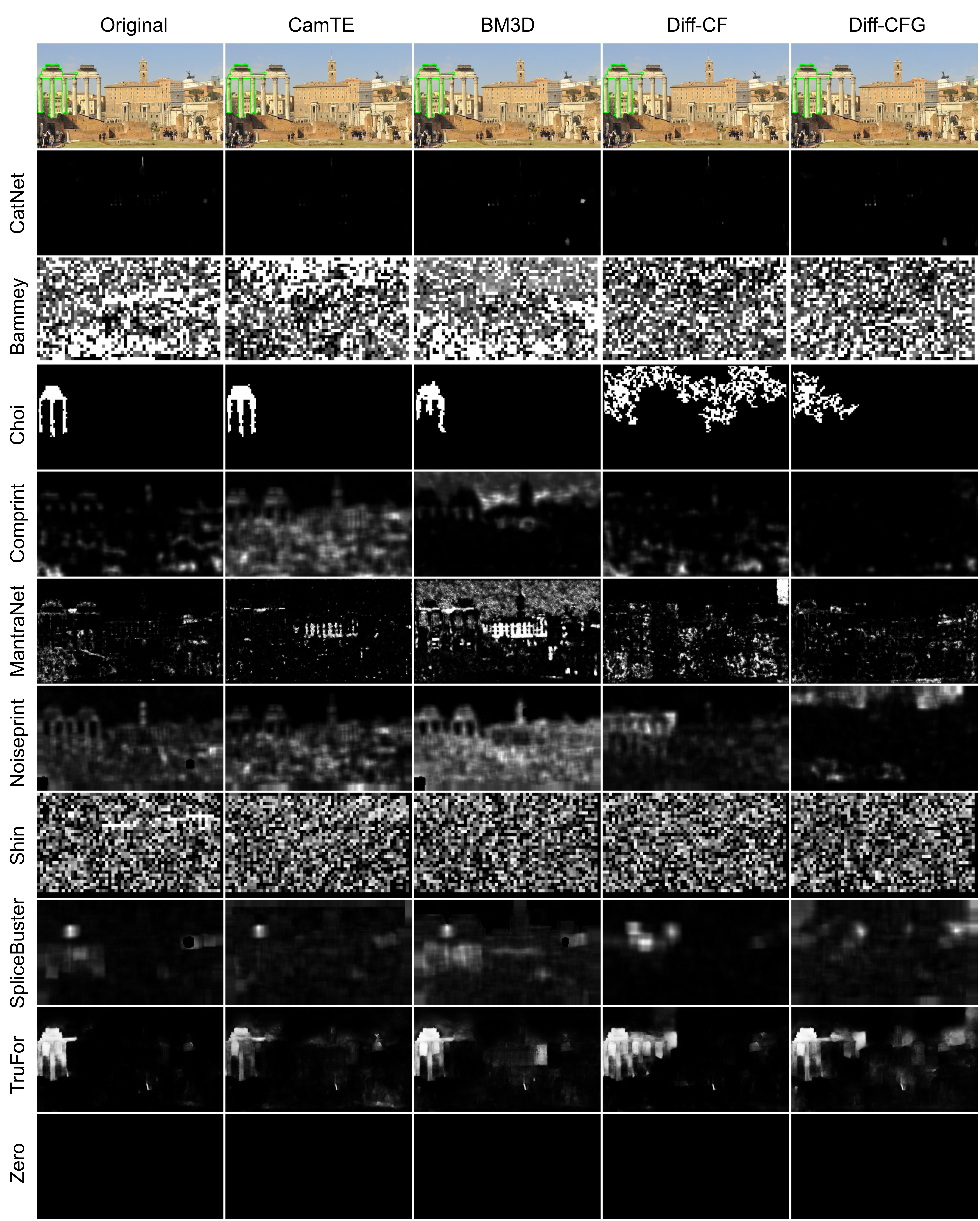}
   \caption{Results obtained by different forensics methods on the different versions of image \texttt{rbc87504ct} from the Korus dataset~\cite{korus1, korus2}. In this image, we observe that only Choi and TruFor detect the forgery in the original image. Choi still detects the forgery once BM3D and CamTE are applied, but is unable to detect it once \emph{Diff-CF} or \emph{Diff-CFG}}
\end{figure*}

\begin{figure*}[t]
  \centering
\includegraphics[width = .9\textwidth]{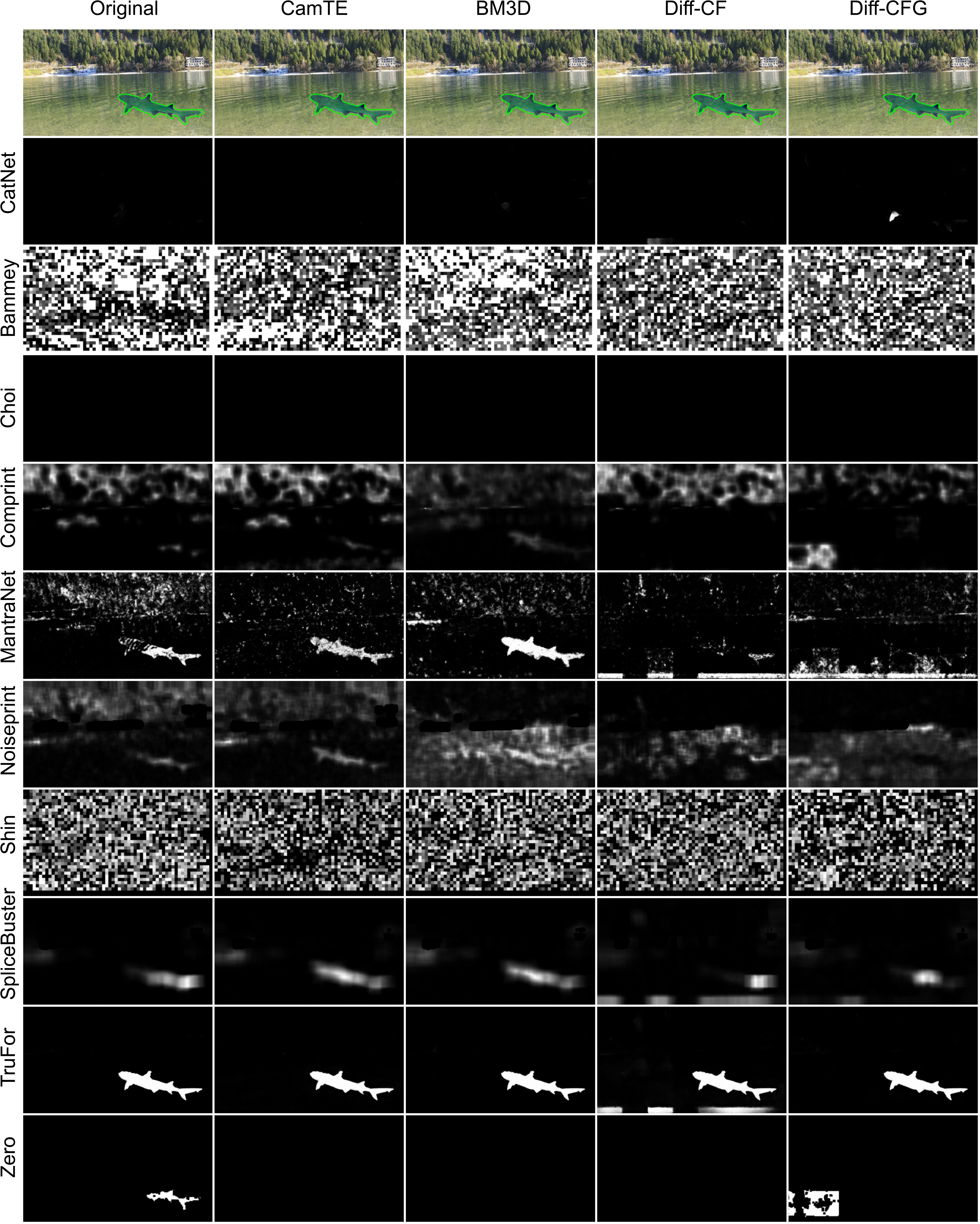}
   \caption{Results obtained by different forensics methods on the different versions of image \texttt{re6d1dc1et} from the Korus dataset~\cite{korus1, korus2}.}
\end{figure*}

\begin{figure*}[t]
  \centering
\includegraphics[width = .8\textwidth]{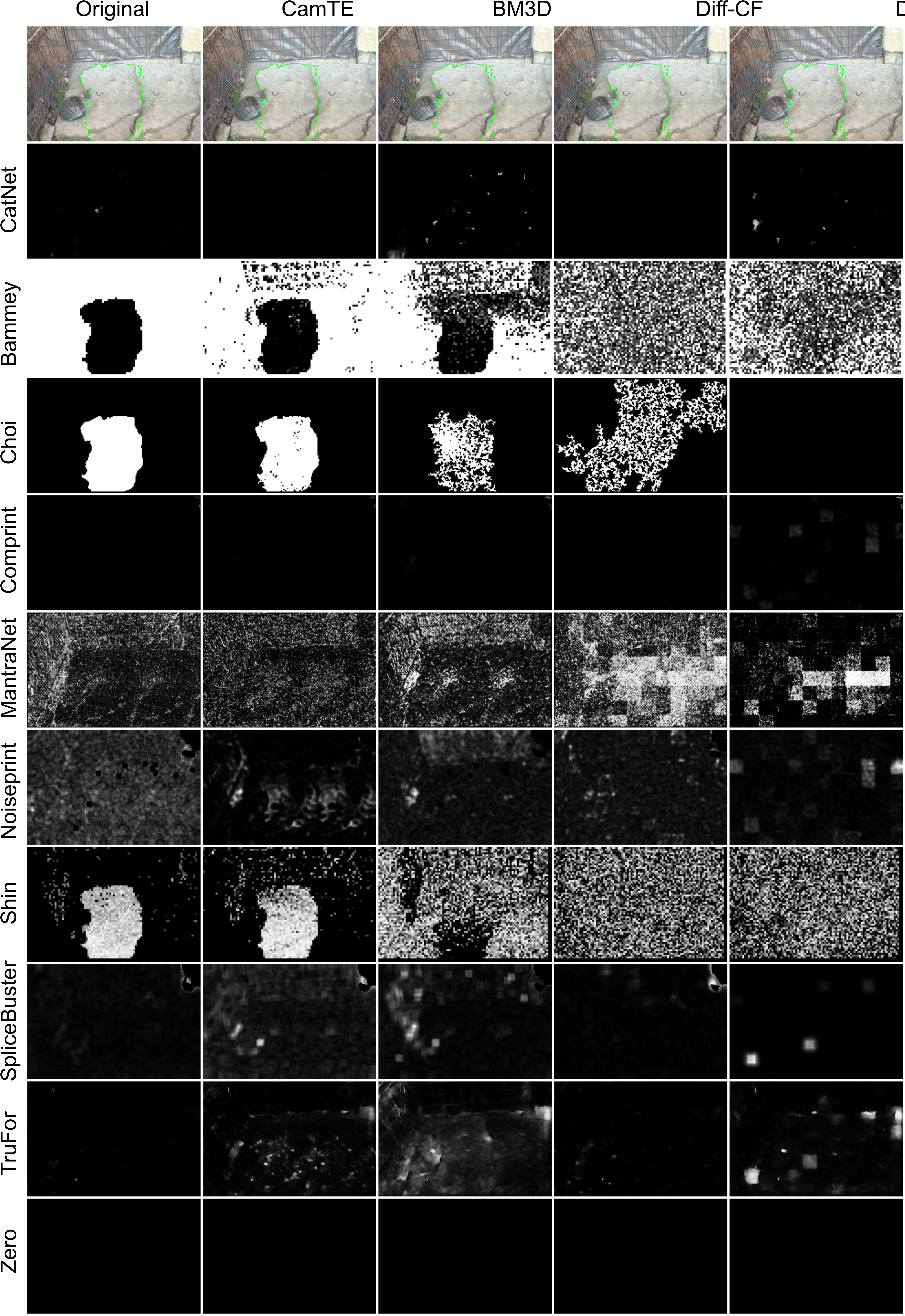}
   \caption{Results obtained by different forensics methods on the different versions of image \texttt{lone\_cat\_copy} from the FAU dataset~\cite{FAU_dataset}.}
\end{figure*}

{\small
\bibliographystyle{ieee_fullname}
\bibliography{egbib}
}